\newcommand {\esiste} {\exists}

\newcommand {\sx} {\langle}
\newcommand {\dx} {\rangle}

\newcommand {\appartiene} {\in}
\newcommand {\emme} {\mathcal{M}}

\newcommand {\tc} {\mid}
\newcommand {\vuoto} {\emptyset}

\newcommand {\diverso} {\neq}

\newcommand{\tip}{{\bf T}}

\newcommand{\alc}{\mathit{ALC}}
\newcommand{\alcrt}{\mathit{ALC}+\tip_{\Ra \ }}

\newcommand{\el}{\mathit{EL}}
\newcommand{\elbot}{\mathit{EL}^{\bot}}
\newcommand{\elPP}{\mathit{EL}^{++}}

\newcommand{\shiq}{\mathit{SHIQ}}

\newcommand{\sroel}{\mathit{SROEL}(\sqcap,\times)}
\newcommand{\sroelrt}{\mathit{SROEL}(\sqcap,\times)^{\Ra}\tip}
\newcommand{\shiqrt}{\mathit{SHIQ}^{\Ra}\tip}
\newcommand{\wARC}{w^{\footnotesize{A \sqsubseteq \exists R.C}}}

\newcommand{\auxARC}{{aux^{\footnotesize{A \sqsubseteq \exists R.C}}}}

\newcommand{\prj}{{\iota}}

\newcommand{\be}{\begin{enumerate}}
\newcommand{\ee}{\end{enumerate}}

\newcommand{\hide}[1]{}

\newtheorem{theorem}{Theorem}
\newtheorem{lemma}{Lemma} 
\newtheorem{proposition}{Proposition}
\newtheorem{definition}{Definition}
\newtheorem{example}{Example}

\def \cases{\left \{\begin{array}{l}}
\def \endcases{\end{array}\right .}

\newcommand {\Ra} {{\bf R}}

\newcommand {\bes} {\begin{description}}
\newcommand{\ens} {\end{description}}

\newcommand {\beq} {\begin{quote}}
\newcommand {\enq} {\end{quote}}
\newcommand {\bit} {\begin{itemize}}
\newcommand {\enit} {\end{itemize}}

\newcommand {\bbox}{\square}

\newcommand{\sroiq}{\mathit{SROIQ}}
\newcommand{\shoiq}{\mathit{SHOIQ}}

\newenvironment{pozz}{\color{black}}{\color{black}}

\documentclass{my_new_tlp}
\usepackage{mathptmx}

\usepackage{times}
\usepackage{undertilde}
\usepackage{graphicx}
\usepackage{latexsym}
\usepackage{graphicx}

\usepackage{color}
\usepackage{amssymb}

\submitted {6 May 2016}
\revised {8 July 2016}
\accepted{22 July 2016 }

\title[ASP for Minimal Entailment in a Rational Extension of SROEL]
        {{ASP for Minimal Entailment \\
        in a Rational Extension of SROEL
        }}

  \author[L.Giordano and D.Theseider Dupr\'e]
         {Laura Giordano and Daniele Theseider Dupr\'e\\
         DISIT - Universit\`a del  Piemonte Orientale, Alessandria, Italy \\
         \email{laura.giordano@uniupo.it, dtd@di.unipmn.it}}

\begin{document}
\bibliographystyle{acmtrans}

\maketitle
\thispagestyle{myheadings}

\begin{abstract}
In this paper we exploit Answer Set Programming  (ASP) for reasoning in a rational extension ${\mathit SROEL} (\sqcap,$ $ \times)^{\Ra \ }\tip$ of the
low complexity description logic $\sroel$, which underlies the OWL EL ontology language.
In the extended language, a typicality operator $\tip$ is allowed to define concepts $\tip(C)$ (typical $C$'s) under a rational semantics.
It has been proven that instance checking under rational entailment has a polynomial complexity.
To strengthen rational entailment,
in this paper we consider a minimal model semantics. We show that, for arbitrary $\sroelrt$ knowledge bases, instance checking
under minimal entailment is $\Pi^P_2$-complete.
Relying on a Small Model result, where models correspond to answer sets of a suitable ASP encoding,
we exploit Answer Set Preferences (and, in particular,  the {\em asprin} framework)
for reasoning under minimal entailment.
The paper is under consideration for acceptance in Theory and Practice of Logic Programming.

\end{abstract}

\section{Introduction}
In the context of work that aims at the convergence of description logics (DLs) and rule-based languages (see, e.g., the invited talk by
Hitzler at ICLP 2013), some
combinations of DLs and LP languages have been proposed,
for instance under the answer set semantics
\cite{Eiter2008}, under the MKNF semantics \cite{KnorrECAI12}, as well as in Datalog +/- \cite{Gottlob14}.
Many extensions of DLs have also been proposed
\cite{Straccia93,baader95b,donini2002,lpar2007,Eiter2008,kesattler,sudafricaniKR,bonattilutz,casinistraccia2010,rosatiacm,KnorrECAI12,CasiniDL2013,AIJ,bonattiAIJ15}
in order to deal with defeasible reasoning,
to allow for prototypical properties of concepts, and to deal with defeasible inheritance.

In this paper we show that a non-trivial form of defeasible reasoning in DLs can be mapped to Answer Set Programming (ASP) \cite{GelfondLeone02}.
In particular, we focus on rational extensions of DLs 
developed along the lines of the preferential semantics introduced by Kraus, Lehmann and Magidor \cite{KrausLehmannMagidor:90,whatdoes}
and, specifically on {\em ranked interpretations}.
These extensions  model 
typical, defeasible, properties of individuals besides strict ones,
extending DLs semantics with a preference relation among domain individuals.
For the logic $\alc$, a preferential extension has been proposed in \cite{lpar2007,FI09},
introducing a typicality operator $\tip$ in the language, which allows defeasible inclusions $\tip(C) \sqsubseteq D$
(``the typical $C$ elements are $D$s") to be expressed.
A rational extension of $\alc$ has been developed in \cite{sudafricaniKR}
allowing defeasible inclusions of the form $C \utilde{\sqsubset} D$, based on ranked interpretations (i.e., modular preferential interpretations).
Preferential description logics have been used as the basis of stronger non-monotonic constructions,
such as the rational closure construction, originally defined by Lehmann and Magidor \shortcite{whatdoes} and developed for $\alc$ in
\cite{Straccia93,CasiniDL2013,AIJ15}.  
In particular, in \cite{AIJ15}  a rational closure construction has been presented which is based on a rational extension of $\alc$
with the typicality operator, and which is characterized semantically by 
the minimal (canonical) rational models of the knowledge base (KB).

In this work we consider a rational extension $\sroelrt$ of the low-complexity description logic $\sroel$
\cite{KrotzschJelia2010},  an extension of $\elPP$  \cite{rifel}, with local reflexivity, conjunction of roles and concept products,
which is at the basis of OWL EL.

It has been shown in \cite{DL2016+CILC} that, in $\sroelrt$, instance checking under rational entailment
can be solved in polynomial time, defining a Datalog translation for normalized knowledge bases which builds on the materialization calculus in \cite{KrotzschJelia2010}.
However, it is widely recognized that rational entailment only allows a rather weak kind of inference,
and minimal and canonical model semantics have been developed to capture stronger non-monotonic inferences \cite{whatdoes}.
We show that the notion of minimal canonical model introduced in \cite{AIJ15} as a semantic characterization of the rational closure for $\alc$ is not adequate to capture some knowledge bases in $\sroelrt$,
and we introduce an alternative minimal model semantics, by weakening the requirement that models have to be canonical,
defining the notions of $\tip$-complete and $\tip$-minimal model of a KB.
We show that, for the KBs for which there are minimal canonical models, all determining the same ranking of concepts as the rational closure,
$\tip$-minimal models capture the same defeasible inferences as  minimal canonical models.

In this paper we exploit ASP for reasoning in the $\tip$-minimal models of a KB.
Exploiting the fact that, in modular preferential interpretations, the preference relation can be equivalently formulated by a rank function,
we provide a Small Rank theorem that ensures that the number of different ranks to be considered in rational models of a KB
can be limited by the number of the concepts ``$\tip(C)$'' occurring in the KB.
Relying on this result, we define an ASP encoding for any normalized $\sroelrt$ KB,
showing that  the {\em answer sets} of the ASP encoding correspond to the ranked models of the KB.
This result also provides a Small Model Theorem for normalized $\sroelrt$ knowledge bases.
The ASP encoding builds 
on the materialization calculus for $\sroel$ presented in \cite{KrotzschJelia2010}.

Reasoning under {\em minimal entailment} requires reasoning on the (possibly multiple) minimal models of a KB.
We show that deciding instance checking under $\tip$-minimal entailment is a  $\Pi^P_2$-complete problem
and we use the ASP encoding of the KB to compute  the answer sets corresponding to $\tip$-minimal models.
In particular, we exploit optimization by multi-shot ASP solving in  the {\em asprin} framework for Answer Set Preferences \cite{BrewkaIJCAI15}.
This approach can be easily adapted to deal with ABox minimization,
by minimizing the ranks of named individuals.
This strictly relates to the rational closure of ABox  in \cite{AIJ15}.

\section{A rational extension of $\sroel$}

In this section
we extend the notion of concept in $\sroel$, defined by Kr\"{o}tzsch \shortcite{KrotzschJelia2010}, %
adding typicality concepts
(we refer to \cite{KrotzschJelia2010} for a detailed description of the syntax and semantics of $\sroel$).
We let ${N_C}$ be a set of concept names, ${N_R}$ a set of role names
  and ${N_I}$ a set of individual names.  
A concept in $\sroel$ is defined as follows:
\begin{center}
 $C:= A \tc \top \tc \bot \tc  C \sqcap C \tc \exists R.C \tc \exists R.Self  \tc \{a\}$

 \end{center}
where $A \in N_C$, $R \in {N_R}$ and $a \in N_I$.
We introduce a notion of {\em extended concept} $C_E$ as follows:
\begin{center}
$C_E:= C \tc \tip(C) \tc C_E\sqcap C_E \tc \exists R.C_E$
 \end{center}
\noindent
where $C$ is a $\sroel$ concept.
Hence, any concept of $\sroel$ is also an extended concept; a typicality concept $\tip(C)$ is an extended concept and can occur
in conjunctions and existential restrictions, but it cannot be nested.

A KB is a triple $\mathit{(TBox, RBox, ABox)}$. $\mathit{TBox}$ contains a finite set
of  {\em general concept inclusions} (GCI) $C \sqsubseteq D$, where $C$ and $D$ are extended concepts;
$\mathit{RBox}$ contains a finite set of {\em role inclusions} of the form $S \sqsubseteq T$,
$R \circ S \sqsubseteq T$,
$S_1 \sqcap S_2 \sqsubseteq T$,
$C \times D \sqsubseteq T$ and
$R \sqsubseteq C \times D$, where $C$ and $D$ are concepts, $R,S,S_1,S_2,T \in N_R$.
$\mathit{ABox}$ contains {\em individual assertions} of the form $C(a)$ and $R(a,b)$, where $a, b \in N_I$, $R \in N_R$ and $C$ is an extended concept.
Restrictions are imposed on the use of roles as in \cite{KrotzschJelia2010}.


Consider the following example of KB, stating that: typical Italians have black hair; typical students are young;
they hate math, unless they are nerd (in which case they love math); all Mary's friends are typical students.
We also assert that  Mary is a student, that Mario is an Italian student and a friend of Mary,
Luigi is a typical Italian student, and Paul is a typical young student.\normalcolor
\begin{example}\label{exa:1}
$\mathit{TBox}$:
$(a) ~ ~  \tip(\mathit{Italian}) \sqsubseteq \mathit{ \exists hasHair.\{Black\}}$ \ \ \
$(b) ~ ~  \tip(\mathit{Student})\sqsubseteq\; \mathit{Young}$ \ \ \

$(c) ~ ~  \tip(\mathit{Student})\sqsubseteq\; \mathit{MathHater}$  \ \ \ \ \ \ \ \ \ \ \ \ \ \ \ \ \ \ \
$(d) ~ ~ \tip(\mathit{NerdStudent})\sqsubseteq\; \mathit{MathLover}$ \ \ \ \ \ \ \ \ \ \ \ \ \

$(e) ~ ~ \mathit{NerdStudent} \sqsubseteq \mathit{Student} $ \ \ \ \ \ \ \ \ \ \ \ \ \ \ \ \ \ \ \ \ \ \ \ \
$(f) ~ ~ \mathit{MathLover} \sqcap \mathit{MathHater} \sqsubseteq \bot$

$(g) ~ ~ \exists \mathit{friendOf}.\{\mathit{mary}\} \sqsubseteq \tip(\mathit{Student})$  \ \ \ \ \ \ \ \
$(h) ~ ~ \mathit{\exists hasHair.\{Black\}} \sqcap \mathit{\exists hasHair.\{Blond\}} \sqsubseteq \bot$

\smallskip
\noindent
$\mathit{ABox}$: 
$~~ \mathit{Student}(\mathit{mary}), ~
\mathit{friendOf}(\mathit{mario}, \mathit{mary}), ~ $
$(\mathit{Student} \sqcap \mathit{Italian})(\mathit{mario}), ~ $
$ \tip (\mathit{Student} \sqcap \mathit{Italian})$ $(luigi), ~ $
$\tip (\mathit{Student} \sqcap \mathit{Young})(paul)  $, $\tip (\mathit{NerdStudent} \sqcap \mathit{Tall})(bob)  $ 

\end{example}

$\tip(C)$ is intended to select the most typical instances of $C$ and
can occur anywhere except from being nested in a $\tip$ operator
(as it can be seen from the semantics below, the operator $\tip$ is idempotent). 
Occurrence of typicality on the r.h.s. of inclusions
can be used, e.g., to state that typical working students inherit properties of typical students
($\tip(\mathit{Student} \sqcap \mathit{Worker}) \sqsubseteq \tip(\mathit{Student})$),
or to state that there are typical Italian students:
$\top \sqsubseteq \exists U. \tip(\mathit{Student} \sqcap \mathit{Italian})$,
where $U$ is the universal role ($\top \times \top \sqsubseteq U$).
As inclusion $\sqsubseteq$  is strict and $\tip(C)$ is a concept, by standard DL inference
we can conclude that Mario is a typical student (by (g)) and young (by (b)).
Moreover, we expect that, according to desired properties of defeasible inclusions,
Paul, who is a typical young student,
inherits the property of typical students of being math haters,
while for Bob the more specific property of typical nerd students of being math lovers should prevail. 

Following  \cite{FI09,AIJ15}, a semantics for the extended language is defined,
adding to interpretations in  $\sroel$ \cite{KrotzschJelia2010} a \emph{preference relation} $<$ on
the domain, which is intended to compare the ``typicality''
of domain elements.
The typical instances of a concept $C$, i.e., the instances of
$\tip(C)$, are the instances of $C$ that are minimal with respect
to $<$. 
As here we consider a rational extension of $\sroel$,  we assume the preference relation $<$
to be modular as in \cite{sudafricaniKR,AIJ15}.

\begin{definition} \label{semalctr}
A $\sroelrt$ interpretation $\emme$ is any
structure $\langle \Delta, <, \cdot^I \rangle$ where:
\begin{itemize}
\item
$ \Delta$ is a domain;
 $\cdot^I$ is an interpretation function that maps each
concept name $A$ to set $A^I \subseteq  \Delta$, each role name $r$
to  a binary relation $R^I \subseteq  \Delta \times  \Delta$,
and each individual name $a$ to an element $a^I \in  \Delta$.
The interpretation function $\cdot^I$ is extended to complex concepts as usual:\\
$\top^I=\Delta$; \ \ \ \ \ \ \ $\bot^I=\emptyset$; \ \ \ \ \ \ \
$\{a\}^I= \{a^I\}$;  \ \ \ \ \ \ \
$(C \sqcap D)^I$= $C^I \cap D^I$; \\
$(\esiste R.C)^I$= $\{x \in \Delta \tc \exists y \in C^I: (x,y) \in R^I \}$;  \ \ \ \ \ \ \
$(\exists R.Self)^I$= $\{x \in \Delta \tc (x,x) \in R^I\}$.

\item
$<$ is an irreflexive, transitive, 
well-founded and modular\footnote{An irreflexive and transitive relation $<$ is  {\em well-founded} if,
for all  $S \subseteq \Delta$, for all $x \in S$, either $x \in Min_<(S)$
or $\exists y \in  Min_<(S)$ such that $y < x$. It is {\em modular} if, for all $x,y,z \in \Delta$, $x <y$ implies $x<z$ or $z<y$.}
relation over $\Delta$.
\item
Let  $Min_<(S)= \{u: u \in S$ and $\nexists z \in S$ s.t. $z < u \}$;
the interpretation of concept $\tip(C)$ is defined as follows:
$(\tip(C))^I = Min_<(C^I)$
\end{itemize}
\end{definition}
As in  \cite{whatdoes}, modularity in preferential models can be equivalently defined by postulating the existence of
a rank function $k_{\emme}: \Delta \longmapsto \Omega$, where $\Omega$ is a totally ordered set.
Hence, modular preferential models are called $\mathit{ranked \ models}$.
The preference relation $<$ can be defined from $k_{\emme}$ as follows:
$x < y$ if and only if $k_{\emme}(x) < k_{\emme}(y)$.
In the following, we assume that a rank function $k_{\emme}$ is always associated with any model $\emme$.
We also define
the {\em rank, $k_{\emme}(C),$ of a concept $C$ in the model $\emme$} as $k_{\emme}(C) = min\{k_{\emme}(x) \tc
x \in C^I\}$ (if $C^I=\vuoto$, then
$C$ has no rank and we write $k_{\emme}(C)=\infty$).
Given 
an interpretation $\emme$ 
the notions of satisfiability and entailment are defined as usual:
\begin{definition}[Satisfiability and rational entailment]\label{Def-ModelSatTBox-ABox}
An interpretation $\emme=\sx\Delta, <, \cdot^I\dx$ satisfies:\\
$\bullet$ \  a concept inclusion $C \sqsubseteq D$ if \  $C^I \subseteq D^I$;\\
$\bullet$ \   a {role inclusion} $S \sqsubseteq T$ if  \ $S^I \subseteq T^I$;\\
$\bullet$ \  a {generalized role inclusion} $R \circ S \sqsubseteq T$ if \  $R^I \circ S^I \subseteq T^I$
	(where $R^I \circ S^I = \{(x,z) \mid (x,y)\in R^I$ and
	\mbox{\ \ \ }$(y,z) \in S^I$, for some $y\in \Delta\}$);\\
$\bullet$ \   a {role conjunction} $S_1 \sqcap S_2 \sqsubseteq T$ if \  $S_1^I \cap S_2^I \subseteq T^I$;\\
$\bullet$ \   a {concept product axiom} $C \times D \sqsubseteq T$  if \ $C^I \times D^I \subseteq T^I$;\\
$\bullet$ \  a {concept product axiom} $R \sqsubseteq C \times D$ if \ $R^I \subseteq C^I \times D^I$;\\
$\bullet$ \   an assertion $C(a)$ if $a^I \in C^I$;\\
$\bullet$ \   an assertion $R(a,b)$ if $(a^I,b^I) \in R^I$.

Given  a KB $\mathit{K=(TBox, RBox, ABox)}$, an interpretation $\emme=$$\sx \Delta, <, \cdot^I \dx$ {\em satisfies}
$\mathit{TBox}$  (resp., $\mathit{RBox}$, $\mathit{ABox}$) if
$\emme$ satisfies   all  axioms in $\mathit{TBox}$ (resp., $\mathit{RBox}$, $\mathit{ABox}$),
and we write $\emme \models \mathit{TBox}$ (resp., $\mathit{RBox}$, $\mathit{ABox}$).
An interpretation $\emme = \sx \Delta, <, \cdot^I \dx$ is a {\em model} of $K$ (and we write $\emme \models K$) if
$\emme$ satisfies   all the axioms in $\mathit{TBox}$, $\mathit{RBox}$ and $\mathit{ABox}$.

Let a query $F$ be either a concept inclusion $C\sqsubseteq D$, where $C$ and $D$ are extended concepts, or an individual assertion.
{\em $F$ is rationally entailed by $K$}, written $K \models_{sroelrt} F$, if for all models $\emme=$$\sx \Delta, <, \cdot^I \dx$ of $K$,
$\emme$ satisfies $F$.
\end{definition}

As shown in \cite{FI09} for the preferential extension of $\alc$, the meaning of $\tip$ can be split into two parts: for any
element $x\in \Delta$,  $x \in (\tip(C))^I$ when
(i) $x \in C^I$, and (ii) there is no $y \in C^I$ such that $y < x$.
The latter can be expressed by introducing a G\"odel-L\"ob
modality $\bbox$ and interpreting the preference
relation $<$ as the accessibility relation of this modality.
Well-foundedness of $<$ ensures that typical elements of $C^I$ exist
whenever $C^I \diverso \vuoto$, avoiding infinitely
descending chains of elements.
The interpretation of $\bbox$ in $\emme$ is as follows:
      $ (\bbox C)^I = \{x \in \Delta \tc $  for every $y \appartiene \Delta$, if
    $y < x$ then $y \in C^I \}.$
\noindent 
The following result, from \cite{FI09}, works as well for typicality based on the rational semantics and for $\sroelrt$,
and will be exploited in Section 4 to define an encoding of $\sroelrt$ in ASP:
\vspace{-0.2 cm}
\begin{proposition}\label{Relation between T an box}
Given a model $\emme$, a concept $C$ and an element $x \in \Delta$:
$x \in (\tip(C))^I \ \mbox{iff}$  $ x \in (C \sqcap \bbox \neg C)^I$
\end{proposition}

In the rest of the paper, we mainly focus on the problem of instance checking.
In particular, 
we propose an inference method in ASP for instance checking in $\sroelrt$ under a minimal model semantics, assuming
the knowledge base is in normal form. 

A {KB in $\sroelrt$ is in {\em normal form} if it admits the axioms of a $\sroel$ KB in normal form:
\begin{center}
$C(a)$ \ \ \ \ \ \ \ $R(a,b)$  \ \ \ \ \ \ \ $A \sqsubseteq \bot$ \ \ \ \ \ \ \ $\top \sqsubseteq C$ \ \ \ \ \ \ \ $A \sqsubseteq \{c\}$  \ \ \ \ \ \ \
$A \sqsubseteq C$ \ \ \ \ \ $A \sqcap B \sqsubseteq C$ \\
$\exists R. A \sqsubseteq C$ \ \ \ \ \ $A \sqsubseteq \exists R. B$ \ \ \ \ \ \ \
$\{a\} \sqsubseteq C$ \ \ \ \ \ $\exists R. \mathit{Self} \sqsubseteq C$ \ \ \ \ \ $A \sqsubseteq \exists R. \mathit{Self}$\\
$R \sqsubseteq T$  \ \ \ \ \ \ \ $R \circ S \sqsubseteq T$  \ \ \ \ \ \ \ $R \sqcap S \sqsubseteq T$  \ \ \ \ \ \ \ $A \times B \sqsubseteq R$  \ \ \ \ \ \ \ $R \sqsubseteq  C \times D $
\end{center}
(where $A,B,C,D \in N_C$, $R,S,T \in N_R$ and $a,b,c \in N_I$)
and, in addition, it admits axioms of the form:
$A\sqsubseteq \tip(B)$\ \  and \ \ $\tip(B) \sqsubseteq C$
with $A,B,C \in N_C$.
Extending the results in \cite{rifel} and in \cite{KrotzschJelia2010}, it is easy to see that, given a $\sroelrt$ KB, a semantically equivalent KB in normal form (over an extended signature) can be computed in linear time.
For details we refer to \cite{DL2016+CILC}, where it is proved that, for normalized $\sroelrt$ KBs, rational entailment can be computed in polynomial time, exploiting a Datalog encoding extending
the materialization calculus for $\sroel$ 
in \cite{KrotzschJelia2010}.

A small rank result can also be proved for $\sroelrt$.
Let $K$ be a knowledge base in $\sroelrt$ 
and let $C_{K}$ be the set of the concepts $C$ such that $\tip(C)$ occurs in $K$.
We prove that, if $K$ is satisfiable, then
there is a model 
of $K$ such that the rank of each element in $\emme'$
is less than the number $max_K$ of concepts in $C_{K}$. 

\begin{theorem}[Small Rank]\label{SmallRank:th}
Let $\mathit{K=(TBox, RBox, ABox)}$
be a normalized $\sroelrt$ knowledge base. Given any
model $\mathcal{M}=(\Delta, <, \cdot^I )$ of $K$, 
there exists a model
$\mathcal{\emme'}=(\Delta, <', \cdot^{I'})$ of $K$ (over the extended language) such that, for all $x \in \Delta$:
(i) $k_{\emme'}(x) \leq max_K$;
(ii) for all $C \in N_C$, $x \in C^{I'}$ iff $x \in C^{I}$; and
(iii) for all $C \in C_{K}$, $x \in (\tip(C))^{I'}$ iff $x \in (\tip(C))^{I}$.
\end{theorem}

The proof can be found in Appendix A.
As a consequence of this result, we can restrict our consideration to models $\emme$ of the KB such that
$k_{\emme}: \Delta \longmapsto \{0 ~ .. ~ max_K \}$.


\section{Minimal entailment}

In Example 1, we cannot conclude using rational entailment 
that all typical young Italians have black hair (and that Luigi has black hair),
as we do not know whether there is some typical Italian who is young.
To support such a stronger nonmonotonic inference, a minimal model semantics can be used to select the interpretations where individuals are as typical as possible.

While restricting to minimal models allows the typicality of domain individuals to be maximised,
some alternative notions of minimality have been considered in the literature \cite{AIJ,CasiniDL2013,AIJ15}. 
In particular, in \cite{AIJ15} a notion of minimality is considered for 
$\alc$ with typicality where models with the same domains and the same interpretations of concepts are compared and
the ones minimizing the ranks of domain elements are preferred.

Namely, an interpretation $\emme = $$\langle \Delta, <, I \rangle$ {\em is preferred to} $\emme' =
\langle \Delta', <', I' \rangle$
($\emme \prec \emme'$) if:
$\Delta = \Delta'$;
 $C^I = C^{I'}$ for all (non-extended) concepts $C$;
for all $x \in \Delta$, 
$ k_{\emme}(x) \leq k_{\emme'}(x)$, and there exists
$y \in \Delta$ such that $ k_{\emme}(y) < k_{\emme'}(y)$.

Given a query $Q$ (where $Q$ can be an assertion $C(a)$ or $\tip(C)(a)$
or an inclusion $C \sqsubseteq D$ or $\tip(C) \sqsubseteq D$)
we say that $Q$ is {\em minimally entailed} by a knowledge base $K$
if $Q$ is satisfied in all the minimal models of $K$.

It has been observed \cite{AIJ15}, that this notion of minimality alone fails to select the intended minimal models.
For instance, consider a $K$ containing the inclusions (c), (d), (e) (f) from Example \ref{exa:1}.
With the above notion of minimality, $\tip(\mathit{NerdStudent \sqcap Tall})\sqsubseteq\; \mathit{MathLover}$ is not entailed by $K$,
i.e. we cannot conclude that all the typical tall nerd students are math lovers (something we would like to conclude, given the irrelevance of being tall
with respect to being nerd students). Indeed, there is a minimal model $\emme$ of $K$ in which a typical tall nerd student is not a math lover,
as there is no tall nerd student which is also a math lover in $\emme$.

The explanation that $\emme$ does not contain sufficiently many individuals
has led to restrict the consideration to models, called {\em canonical}, that include
a domain individual for any set of concepts $\{ C_1,\ldots,C_n\}$ consistent with the KB
(where the $C_i$'s are non-extended concepts occurring in KB or their negations).
For $\alc$ and $\shiq$ it has been shown \cite{AIJ15,DL2014} that minimal canonical models provide
a semantic characterization of the rational closure 
of TBox which, however, is defined only for KBs where typicality concepts only occur on the l.h.s.\ of inclusions (we call them {\em simple} KBs).
This holds in particular for $\el^\bot$ plus typicality (which is a fragment of $\alc$).
In the general case, a KB in $\sroelrt$ may have multiple minimal models with incomparable ranking functions. Consider the following example:

\begin{example}
Let $K$ be a knowledge base such that:
$\mathit{RBox}=\{\mathit{C \times D \sqsubseteq R}\}$,
$\mathit{ABox}=\emptyset$,
and $\mathit{TBox}$ contains the inclusions
(1) \ $C \sqcap D \sqsubseteq \bot$, \
(2) \ $\tip(\top) \sqcap  \exists R. \tip(\top) \sqsubseteq \bot$, \
(3) \ $\tip(C) \sqsubseteq  E$, \
(3) \ $\tip(D) \sqsubseteq  E$. \
Observe that, by the $RBox$ inclusion, each $C$ element is in relation $R$ with all $D$ elements and,
by inclusion (2) in $TBox$, it is not the case that two elements of rank 0 (the rank of typical $\top$ elements)
can be in the relation $R$.
So, it is not possible that a $C$ element and a $D$ element have both rank $0$
and, in all minimal canonical models,  either $C$ has rank $0$ and $D$ has rank $1$, or vice-versa.

\end{example}
The existence of alternative minimal models for a KB with free occurrences of typicality
was observed in \cite{Booth15} for Propositional Typicality logic (PTL), a propositional language with negation.
While the existence of alternative minimal canonical models is not per se a problem, it may happen that  a KB in $\sroelrt$ {\em has no canonical model}
at all. This problem was already pointed out for expressive logics such as $\shoiq$ \cite{DL2014}.
For instance, if a KB contains the inclusion $\mathit{ \{bob\} \sqcap Student  \sqcap Worker \sqsubseteq \bot}$,
it cannot have a canonical model. In fact, while the two sets of concepts $\{\mathit{ \{bob\}, Student} \}$ and $\{\mathit{ \{bob\}, Worker} \}$
are both consistent with the KB, there is no canonical model which contains
an instance of $\mathit{ \{bob\} \sqcap Student}$ and
one of $\mathit{ \{bob\} \sqcap Worker}$  (as {\em bob} can be a student or a worker, but not both).

Examples like this one suggest that an alternative requirement to the canonical model condition would be needed to extend
the minimal model semantics to a larger set of $\sroelrt$ KBs.
In essence, the canonical model condition requires that a model must contain instances of all (the sets of) concepts occurring in the KB that are consistent with it.
This condition can be weakened by requiring that {\em only for the concepts $C$ such that $\tip(C)$ occurs in the KB $K$ (or in the query)},
an instance of $C$ is required to exist in the model, when $C$ is satisfiable in $K$} (i.e., if there is a model $\emme'$ of $K$ such that $C^{I'} \neq \emptyset$).
We call such models {\em $\tip$-complete}. Let $K$ be a KB and $Q$ a query.
Let ${\cal T}_{K,Q}$= $\{ C \; \mid \; \tip(C)$ occurs in $K$ or in $Q$  and $C$ is satisfiable in $K$\}.
When the query has the form $\tip(C) \sqsubseteq D$, ${\cal T}_{K,Q}$ also includes the two concepts $C \sqcap D$ and $C \sqcap \neg D$ 
when satisfiable in $K$.
\vspace{-0.1 cm}
\begin{definition} \label{Tcomplete}
A model $\emme$ is {\em $\tip$-complete} (wrt $K$, $Q$) if, for all $C \in {\cal T}_{K,Q}$,  $C^I\neq \emptyset$.
\end{definition}

Among $\tip$-complete models, we select the minimal ones according to the following {\em preference relation $\prec_{\tip}$ over the set of ranked interpretations}.
An interpretation $\emme = $$\langle \Delta, <, I \rangle$ {\em is preferred to} $\emme' =
\langle \Delta', <', I' \rangle$ (wrt $K$, $Q$),
written $\emme \prec_{\tip} \emme'$, if, for all $C \in {\cal T}_{K,Q}$, \
$ k_{\emme}(C) \leq k_{\emme'}(C)$, and there exists
$D \in {\cal T}_{K,Q}$ such that $ k_{\emme}(D) < k_{\emme'}(D)$.

\vspace{-0.1 cm}
\begin{definition}
\label{minimal-model}
$\emme$ is a {\em $\tip$-minimal model} of $K$
if it is a $\tip$-complete model of $K$ (wrt $Q$) and it is minimal among  the $\tip$-complete models of $K$  
wrt the preference relation  $\prec_{\tip}$  (wrt $Q$).
\end{definition}

\vspace{-0.2 cm}
\begin{definition}[$\tip$-minimal entailment]\label{minimal-entailment}
Given  a knowledge base $K$ in $\sroelrt$, a query
{\em $Q$ is $\tip$-minimally entailed by $K$}, written $K \models_{\tip min} Q$, if, for all $\tip$-minimal models $\emme$ 
of $K$ (wrt $Q$), $\emme$ satisfies $Q$.
\end{definition}
It can be proved that there is a correspondence between
$\tip$-minimal models and minimal canonical models  for 
knowledge bases $K$ such that:
(i) a canonical model of $K$ exists and
(ii) the ranking $K_{\emme}$ of each canonical model $\emme$ of $K$ is the same as the one determined by the Rational Closure construction.
Let $\models_{min}$ be the minimal entailment based on the minimal canonical models semantics \cite{AIJ15}.

\vspace{-0.1 cm}
\begin{theorem}\label{T_minimality}
Let $K$ be a knowledge base satisfying conditions (i) and (ii) above and $Q$  an inclusion $\tip(C) \sqsubseteq D$
(where $C$ and $D$ are non extended concepts).
Then, $K \models_{\tip min} \tip(C) \sqsubseteq D$ iff $K \models_{min} \tip(C) \sqsubseteq D$.
\end{theorem}
The proof can be found in Appendix A.
In particular, the $\tip$-minimal models semantic and the minimal canonical models semantic coincide
for simple KBs in the intersection of $\alcrt$ and $\sroelrt$ (i.e., in $\el^\bot$ plus $\tip$). 
For this fragment minimal canonical models provide a semantic characterization of rational closure of simple KBs \cite{AIJ15},
so that conditions (i) and (ii) hold.
In addition, $\tip$-minimal models can be defined also for KBs for which no canonical model exists
(for instance,  the KB in Example 1 has a unique $\tip$-minimal model).
In particular, the presence in a KB of an inclusion
$\mathit{ \{bob\} \sqcap Student  \sqcap Worker \sqsubseteq \bot}$,
does not cause the KB to have no $\tip$-minimal models, unless the KB contains other inclusions such as, for instance,
$\mathit{\tip( \{bob\} \sqcap Student) \sqsubseteq E}$ and
$\mathit{\tip( \{bob\} \sqcap Worker)} \sqsubseteq F$,
which would require a $\tip$-complete model to contain  instances of  $\mathit{ \{bob\} \sqcap}$ $\mathit{Student}$
and of  $\mathit{ \{bob\} \sqcap Worker}$, which is not possible.

In Section \ref{Sec:ASPencoding} we show that for a normalized $\sroelrt$ KB we can restrict our attention to small models, whose size is linear in the KB size,
and that we can generate such models as the answer sets of an ASP encoding of the KB.
In Section \ref{sec:minimal_entailment} we introduce a notion of preference among answer sets, to define minimal  $\tip$-complete answer sets of the KB. 
The following result, proved in Appendix B, provides a lower bound on the complexity of $\tip$-minimal entailment:
\vspace{-0.1 cm}
\begin{theorem}\label{th:lowerBound}
Instance checking  in $\sroelrt$ under $\tip$-minimal model semantics is $\Pi^P_2$-hard.
\end{theorem}
While we have introduced the $\tip$-minimal model semantics to capture the minimization of the rank of concepts,
the $\tip$-minimal semantics can be extended as well to maximize the typicality of named individuals. Indeed, in Example 1
we cannot conclude that Mary is a typical student and hence she hates math, unless we assume that Mary is as typical as possible
by preferring those models in which named individuals have the lowest rank. A new notion of preference between models
can indeed be defined by reformulating, for the $\tip$-minimal semantics, the {\em preference wrt ABox} in \cite{AIJ15}
(Def. 26), i.e., by selecting among $\tip$-minimal models those which assign the lowest rank to individual names.

We define a preference $\prec_{ABox}$ between $\tip$-minimal models, as follows.
Let $N_{I,K}$ be the named individuals occurring in $K$
and let  $\emme = $$\langle \Delta, <, I \rangle$ and $\emme' =
\langle \Delta', <', I' \rangle$ be two $\tip$-minimal models of $K$  (wrt $K$, $Q$).
We have that 
$\emme \prec_{ABox} \emme'$, if, for all $a \in N_{I,K}$, \
$ k_{\emme}(a^I) \leq k_{\emme'}(a^I)$, and there exists
$b \in N_{I,K}$ such that $ k_{\emme}(b^I) < k_{\emme'}(b^I)$.
We call $\prec_{ABox}$-minimal the $\tip$-minimal models that have no $\prec_{ABox}$-preferred $\tip$-minimal model.

It is easy to see that also simple KBs satisfying conditions (i) and (ii) of Theorem \ref{T_minimality}, having a unique minimal ranking assignment to concepts,
may have multiple minimal ranking for named individuals.
Consider the following reformulation in $\sroelrt$ of an example dealing with the rational closure of ABox in $\alcrt$ from \cite{AIJ15}.
The reformulation is actually in the fragment $\el^\bot$ plus typicality.

\vspace{-0.1cm}
\begin{example}\label{example-ABox-algorithm-multiple-ranks}
Normally computer science courses ($CS$) are taught by academics ($A$), whereas business courses ($B$) are normally taught by consultants ($C$), while consultants and academics are disjoint, i.e., we have $TBox = \{$
$\exists is\_Teacher\_of. \tip (CS) \sqsubseteq A, ~$
$\exists is\_Teacher\_of. \tip (B) \sqsubseteq C, ~$
$C \sqcap A \sqsubseteq \bot \}$,
$ABox=\{\mathit{ CS(c1), B(c2), is\_Teacher\_of(joe,c1), is\_Teacher\_of(joe,c2)}\}$ and $RBox=\emptyset$.
In the $\tip$-minimal models of the KB, all atomic concepts have rank $0$.
Observe, however, that there is no $\tip$-minimal model in which both $c1^I$ and $c2^I$ have rank $0$,
otherwise, $joe$ would be a teacher of both a typical computer science course and a typical business course, hence he would be both
an academic and a consultant, which is inconsistent.
In the $\prec_{ABox}$-minimal models of $K$ either $c1^I$ has rank $0$ and $c2^I$ has rank $1$, or vice-versa.
\end{example}


\section{Models as answer sets}\label{Sec:ASPencoding}

We map a normalized $\sroelrt$ KB to an ASP program, extending the calculus by Kr\"{o}tzsch \shortcite{KrotzschJelia2010}
with a set of predicates to record the ranks of domain elements as well as the minimal ranks 
for concepts in a ranked model,
thus providing the interpretation of typicality concepts in the model.
Alternative models of the KB, with different rank assignments, correspond to alternative answer sets of the ASP program.
In particular, we show that if the KB has a model $\emme$, then there is an answer set corresponding to a small model of the KB,
which preserves the relative ranks of the concepts in ${\cal T}_{K,Q}$ (according to the small rank result above).

We show that a small number of auxiliary constants (namely, one constant $aux_C$ for each concept $\tip(C)$ occurring in the knowledge case)
need to be introduced in the ASP program, besides the auxiliary constants $aux^{A\sqsubseteq \exists R.C}$ used by the calculus in
\cite{KrotzschJelia2010} to deal with existential restriction.
Generation of (small) models of the KB provides the basis for computing minimal models, and then minimal entailment.
We can show that, in order to reason with minimal entailment, we can restrict, without loss of generality, to models over a domain
containing named individuals plus the auxiliary constants, i.e. to the domain of the models of the ASP encoding.

In this section, we consider the problem of verifying whether, for a given normalized KB, there is a model of the KB satisfying a query
of the form $\tip(C)(a)$ or $C(a)$ with $C \in N_C$.
In Section \ref{sec:minimal_entailment} we address minimal entailment.

Given a normalized knowledge base $K$,
we define $\Pi(K)$, the ASP program associated with $K$, as the union of the following components:
\vspace{-0.2 cm}
\begin{enumerate}
\item
$\Pi_{K}$, the representation of $K$ in ASP, which is based on the input translation in \cite{KrotzschJelia2010}
of a $\sroel$  KB in normal form,
with minor additions for the extended syntax of $\sroelrt$;
\item
$\Pi_{IR}$, the inference rules in \cite{KrotzschJelia2010}, and
additional inference rules for the extended syntax of inclusions with $\tip(C)$ concepts;
\item
$\Pi_{T}$, containing rules and constraints to enforce the $\sroelrt$ semantics;
\end{enumerate}
\vspace{-0.1 cm}

\noindent
{\bf Part 1.}
$\Pi_{K}$ is the representation of $K$ in ASP according to rules that include the ones in \cite{KrotzschJelia2010}, where, to keep a DL-like notation, we do not follow the ASP convention where variable names start with uppercase;
in particular, $A$, $C$, and $R$, are intended as ASP constants corresponding to the same class/role names in $K$. In this representation,
$\mathit{nom(a)}$, $\mathit{cls(A)}$, $\mathit{rol(R)}$ are used for
$\mathit{a \in N_I}$ , $\mathit{A \in N_C}$, $\mathit{R \in N_R}$, and, for example
(the complete set of rules from \cite{KrotzschJelia2010} is reported in Appendix C):
\begin{itemize}
\item
$\mathit{subClass(a,C)}$, $\mathit{subClass(A,c)}$, $\mathit{subClass(A,C)}$ are used for $C(a)$, $A \sqsubseteq \{c\}$, $A \sqsubseteq C$;
\item
$\mathit{supEx(A,R,B,aux_i)}$ is used for $ \mathit{ A \sqsubseteq \exists R . B   }$;
\end{itemize}

In the translation of $ \mathit{ A \sqsubseteq \exists R . B   }$,
$\mathit{aux_i}$  is a new constant, different for each axiom of this form.
The ASP program identifies such names with a fact
$\mathit{auxsupex(aux_i)}$.
The additional mapping for the extended syntax of the $\sroelrt$ normal form is:
\begin{center}
$ \mathit{A\sqsubseteq T(B)} $   $\mapsto  \mathit{supTyp(A,B)} $ \ \ \ \ \ \ \ \
$ \mathit{T(B) \sqsubseteq C} $  $\mapsto  \mathit{subTyp(B,C)} $
\end{center}
Also, we need to add $\mathit{top(\top)}$ to the input specification; moreover,
for any 
concept $C$ occurring in $K$, 
the program includes a fact
$\mathit{auxtc(aux_C,C)}$
where $\mathit{aux_C}$ is a new constant, used in the following as a (name of) a representative typical $C$, in case
$C$ is non-empty.

\smallskip
\noindent
{\bf Part 2.}
$\Pi_{IR}$ contains, with a small variant, the
inference rules in \cite{KrotzschJelia2010} (see rules (1-29) in Appendix C),
for example:
\vspace{-0.1cm}
\begin{tabbing}
$\mathit{inst(x,x) \leftarrow nom(x) } $ \\
$\mathit{inst(x,z) \leftarrow subClass(y,z),inst(x,y)} $ \\
$\mathit{inst(x,z) \leftarrow subEx(v,y,z),triple(x,v,x'),inst(x',y)} $
\end{tabbing}
Note that $inst(c,d)$
for $c,d \in N_I$ means \cite{jeliaReport} that $\{c\} \sqsubseteq \{d\}$, i.e., $c$ and $d$
represent the same domain element.
$\Pi_{IR}$ contains additional inference rules for inclusions with extended concepts:
\vspace{-0.1cm}
\begin{tabbing}
$(30) ~ \mathit{typ(x,z) \leftarrow supTyp(y,z),inst(x,y) } $ \\
$(31) ~ \mathit{inst(x,z) \leftarrow subTyp(y,z),typ(x,y)} $
\end{tabbing}

\noindent
{\bf Part 3.}
$\Pi_{T}$, i.e.\ the set
of rules and constraints to enforce the $\sroelrt$ semantics,
is as follows.
The rules and constraint (where $h,j,k,k1,n$ are  ASP variables, as well as $aux_y$ used in the next
group of rules):
\vspace{-0.5cm}
\begin{tabbing}
$(10)$ \= \kill \\
$(32) ~ \mathit{ind(X) \leftarrow nom(X) } $ \\
$(33) ~ \mathit{ind(X) \leftarrow auxsupex(X)	} $ \\
$(34) ~ \mathit{ind(X) \leftarrow auxtc(X,C)	} $  \\
$(35) ~ \mathit{possrank(0..n) \leftarrow upperbound(n) } $ \\
$(36) ~ \mathit{rank(x,k) \leftarrow ind(x),possrank(k), not ~ hasdiffrank(x,k) } $ \\
$(37) ~ \mathit{hasdiffrank(x,k) \leftarrow possrank(k),rank(x,j), j!=k } $ \\
$(38) ~\mathit{some\_at(k) \leftarrow rank(x,k) } $ \\
$(39) ~ \mathit{\leftarrow some\_at(k1), k1=k+1, possrank(k),  not ~ some\_at(k) } $
\end{tabbing}
define (32-34) the extended set of individual names;
assign (35-37) to each individual name a rank between $0$ and $n$, where $n$ is the number (asserted as
$\mathit{upperbound(n)}$), of $\tip(C)$ concepts in the $KB$ and the query;
without loss of generality, state (38-39) that if no individual has rank $k$, no other individual has rank $k+1$ (and
then, any $h>k$);
this is useful to reduce combinations of rank assignments in case less than $n+1$ different ranks can be used.

The following constraints and rules rely on the correspondence in Proposition \ref{Relation between T an box}
between $\tip(C)$ and $(C \sqcap \bbox \neg C)$, and, using $\mathit{box\_neg(k,C)}$
to represent that $\bbox \neg C$ holds for individuals at rank $k$, relate it to membership of individuals to $\tip(C)$
and to the semantics of typical instances as maximally preferred instances of a concept:
\vspace{-0.5cm}
\begin{tabbing}
$(10)$ \= \kill \\
$(40) ~ \mathit{\leftarrow -box\_neg(k,y), auxtc(aux_y,y), rank(aux_y,h), k \leq h } $ \\
$(41) ~ \mathit{box\_neg(k1,y) \leftarrow box\_neg(k,y),possrank(k1), k1=k-1 } $ \\
$(42) ~ \mathit{-inst(x,y) \leftarrow box\_neg(k,y), rank(x,k1),  k1=k-1 } $ \\
$(43) ~ \mathit{-box\_neg(k1,y) \leftarrow auxtc(aux_y,y),rank(aux_y,k),  inst(aux_y,y),k1=k+1 } $ \\
$(44) ~ \mathit{-box\_neg(k1,y) \leftarrow -box\_neg(k,y), possrank(k1),k1=k+1 } $ \\
$(45) ~ \mathit{box\_neg(n,y) \leftarrow auxtc(aux_y,y),  - inst(aux_y,y),upperbound(n) } $ \\
$(46) ~ \mathit{rank(y,h) \leftarrow nom(y), inst(x,y), rank(x,h) } $ \\
$(47) ~ \mathit{inst(x,y) \leftarrow typ(x,y) } $ \\
$(48) ~ \mathit{typ(x,y) \leftarrow inst(x,y),rank(x,k),box\_neg(k,y) } $ \\
$(49) ~ \mathit{box\_neg(k,y) \leftarrow typ(x,y), rank(x,k) } $ \\
$(50) ~ \mathit{box\_neg(k,y) \leftarrow auxtc(aux_y,y), rank(aux_y,k) } $ \\
$(51) ~ \mathit{inst(aux_y,y) \leftarrow auxtc(aux_y,y), inst(x,y) } $ \\
$(52) ~ \mathit{-inst(aux_y,y) \leftarrow auxtc(aux_y,y), not ~ inst(aux_y,y) } $\\
$(53) ~ \mathit{inst(aux_y,y) \leftarrow auxtc(aux_y,y), not ~ -inst(aux_y,y) } $\\
$(54) ~ \mathit{ \leftarrow bot(z), inst(u, z)} $
\normalcolor
\end{tabbing}
Note that rules (35-37) assign a rank also to the additional individuals $aux_C$.
The constraint (40) states that if an $aux_C$ has rank $h$, $\neg \bbox \neg C$ can only hold at ranks $> h$;
rule (41) states that if $\bbox \neg C$ holds at some rank, it also holds at lower ranks, where (due to rule 42)
individuals are not instances of $C$. Rule (43) states that if  $aux_C$ has rank $k$, and it is indeed
an instance of $C$, then $\neg \bbox \neg C$ holds at $k+1$, and (rule 44) at higher ranks.
Rule (45) is for the case where $aux_C$ is not an instance of $C$; in this case, all domain
elements are not $C$ elements and $\bbox \neg C$ holds for elements at the highest rank (and then at all ranks).

The remaining rules state that:
(46) the same rank is assigned to constants representing the same individual;
(47) typical members of a concept are members;
(48) if $\bbox \neg C$ holds at $k$, instances of $C$ at rank $k$ are typical instances;
(49) if there is a typical instance at rank $k$, $\bbox \neg C$ holds at $k$;
(50) $\bbox \neg C$ holds at the rank of $aux_C$;
(51) $aux_C$ is an instance of $C$ if there is an (other) instance;
(52) and (53) allow to assume that $aux_C$ is either an instance of $C$ or not, in case there are no other instances.
Rule (54) removes answer sets in which the concept $\bot$ has an instance.

\medskip

The representation $\pi_Q$ of a query $Q$ of the form $\tip(C)(a)$ or $C(a)$ (with $C \in N_C$)
is as follows: for a query $Q$ of the form $\tip(C)(a)$,
$\pi_Q$ is  $ typ(a,C)$;
if $Q$ is of the form $C(a)$,
$\pi_Q$ is $inst(a,C))$.
If $Q$ is $\tip(C)(a)$, then $\mathit{auxtc(aux_C,C)}$
is assumed to be in $\Pi (K)$.

We establish a correspondence between models of a knowledge base $K$ falsifying a query $Q$
and answer sets of $\Pi(K) \cup \{- \pi_Q\}$, i.e., the answer sets of $\Pi(K)$ not containing $\pi_Q$.
First we show that answer sets of $\Pi(K) \cup \{- \pi_Q\}$ correspond to models of $K$ falsifying $Q$.

\begin{proposition} \label{AS to models}
Given a knowledge base $K$ in normal form and a query $Q$, if there is an answer set $S$ of the ASP program $\Pi(K) \cup \{- \pi_Q\}$,
then there is a model $\emme$ 
of $K$  such that  $Q$ is  not satisfied in $\emme$.
\end{proposition}

The next proposition shows that if there is a model of $K$ falsifying a query, then there exists
an answer set of $\Pi(K) \cup \{- \pi_Q\}$.
As, by Proposition \ref{AS to models}, such an answer set corresponds to a small model of $K$,
Propositions  \ref{AS to models} and  \ref{models to AS} together provide a small model result for $\sroelrt$.
Their proofs can be found in Appendix D.

\begin{proposition} \label{models to AS}
For a  $\sroelrt$ knowledge base $K$ in normal form and a query $Q$, if $\emme$ 
is a model of $K$
falsifying a query $Q$,
then there exists an answer set $S$ of the ASP program $\Pi(K) \cup \{- \pi_Q\}$.
\end{proposition}

\section{Computing minimal entailment} \label{sec:minimal_entailment}

The $\tip$-minimality condition on models can be reformulated for the answer sets of the ASP encoding.
For a knowledge base $K$ and a query $Q$ of the form $C(a)$ or $\tip(C)(a)$,
we let $Aux_{K,Q}=\{ aux_C \mid \tip(C)$ occurs in $K$ or $Q \}$.
\begin{definition}
An answer set $S$ of $\Pi(K)$ is {\em $\tip$-complete} wrt $K,Q$ if $\mathit{inst(aux_C,C) \in S}$ for all concepts $C$ satisfiable in $K$ and such that $aux_C \in Aux_{K,Q}$.

\noindent
Given two answer sets $S_1$ and $S_2$  of $\Pi(K)$,
$S_1 \preceq_{\tip} S_2$ wrt $K,Q$
if, for all $aux_C \in Aux_{K,Q}$:

(a) if $\{\mathit{rank(aux_C,h_1)}, \mathit{inst(aux_C,C) \} \subseteq S_1}$ and $\mathit{rank(aux_C,h_2) \in S_2}$,
then $h_1 \leq h_2$;

(b) if $\mathit{inst(aux_C,C)  \not\in S_1}$, then $\mathit{inst(aux_C,C)  \not\in S_2}$.

\noindent
An answer set $S$ of of $\Pi(K)$ {\em $S$ is ${\tip}$-minimal}  wrt $K,Q$
if $S$ is minimal, for $\preceq_{\tip}$ wrt $K,Q$, among the answer sets of $\Pi(K)$ which are $\tip$-complete wrt $K,Q$.
\end{definition}
In the definition of $\preceq_{\tip}$, note that (b) always holds for {\em $\tip$-complete} answer sets.
It is easy to see (using Propositions \ref{AS to models} and \ref{models to AS})
that for any $\tip$-minimal model of $K$ falsifying $Q$
there is a  ${\tip}$-minimal answer set of $\Pi(K)$ not containing $\pi_Q$, and vice-versa (see Appendix E,
Proposition 5).
Then $K \models_{\tip min} Q$ if and only if $\pi_Q$ is in all the ${\tip}$-minimal answer sets of $\Pi(K)$ wrt $K,Q$.

In order to make the answer sets of the encoding $\tip$-complete wrt $K,Q$, the following rules:
\vspace{-0.5cm}
\begin{tabbing}
$(10)$ \= \kill \\
$ (55) ~ \mathit{inst(x,y) \leftarrow occurs(y), auxtc(x,y), satisfiable(y) }  $\\
$ (56) ~ \mathit{satisfiable(y) \leftarrow occurs(y), cls(y), not\; unsatisfiable(y) }  $\\
$ (57) ~ \mathit{unsatisfiable(y) \leftarrow occurs(y), cls(y), cls(z), inst\_s(x,z,y), bot(z) }  $\\
$ (58) ~ \mathit{inst\_s(y,y,y) \leftarrow occurs(y) }  $
\end{tabbing}
are added, and a fact $\mathit{occurs(c)}$ is asserted for all concepts $C$ such that $aux_C \in Aux_{K,Q}$.
Rule (55), for all such $Cs$,
makes the auxiliary constant, representative of typical $C$'s, indeed an instance
of $C$, in case $C$ is satisfiable.
Satisfiability is verified using, as done in \cite{KrotzschJelia2010} for subsumption checking,
a version of the basic calculus with an additional parameter.
In rule (57), predicate $\mathit{inst\_s}$ is a version of $\mathit{inst}$
where the third parameter, a concept name $y$,
represents the assumption that the concept is not empty; as in \cite{KrotzschJelia2010},
the name of the concept itself is used for a hypothetical instance of the concept, and
rule (58) provides this membership. Rule (57) then concludes that a concept is not satisfiable if
assuming its non-emptiness leads to infer that $\bot$ has some instance.

The basic calculus, which is extended with the extra parameter,
is, in our case, the Datalog calculus for rational entailment showing
that instance checking under $\models_{sroelrt}$ can be performed
in polynomial time \cite{DL2016+CILC}.
Such a calculus includes the basic calculus in \cite{KrotzschJelia2010} (see Appendix C),
and a set $\Pi_{RT}$ of rules to deal with typicality,
using $typ(a,C)$ to represent $\tip(C)(a)$ as in section \ref{Sec:ASPencoding},
and  including rules (30-31);
however, unlike $\Pi_{T}$ in section \ref{Sec:ASPencoding}, rules in $\Pi_{RT}$ do not assign a rank to each individual, only using predicates
$\mathit{leqRank(x,y),sameRank(x,y)}$ to constrain the ranks of two individuals.
The extra parameter is added as follows: in all rules, $\mathit{occurs(q)}$ is added to the antecedent;
in all literals for predicates $\mathit{inst,triple,self,occurs,typ,leqRank,}$  $\mathit{sameRank}$,
predicate names are replaced with
$\mathit{inst\_s,triple\_s,}$
$\mathit{self\_s,occurs\_s,typ\_s,leqRank\_s,}$
$\mathit{sameRank\_s}$,
and $q$ is added as last parameter.

\medskip

The $\tip$-minimal answer sets are computed using the {\em asprin} framework \cite{BrewkaIJCAI15}
for Answer Set Preferences, which uses multi-shot ASP solving. The framework
allows a user to specify preferences, also using a library of preferences, including
ways for composing basic preferences.
The $\tip$-minimal answer sets
can be selected adding a preference specification that relies on such
a library and is composed of a statement:
\begin{tabbing}
$ \mathit{\#preference(p_i,less(weight)) \{  X,X :: rank(aux_i,X) : possrank(X) \} } $
\end{tabbing}
for each  $\mathit{aux_i} \in Aux_{K,Q}$, that defines a preference, named $\mathit{p_i}$,
for a smaller rank of $\mathit{aux_i}$;
and the statements:
\begin{tabbing}
$ \mathit{ \#preference(p\mbox{-}tbox,pareto) \{ name(p_1) ; \ldots ; name(p_n) \} } $ \\
$ \mathit{ \#optimize(p\mbox{-}tbox) } $
\end{tabbing}
which require an optimal solution with respect to the preference defined as the {\em pareto} combination
of the preferences $\mathit{p_i}$\footnote{Such statements also minimize the rank of an $\mathit{aux_i}$ whose
corresponding concept is not satisfiable, but this is irrelevant;
such a constant will not be instance of any concept, then any rank can be assigned to it.}.
Then, given $\Pi_{\tip min}(K,Q)$ , which is $\Pi(K)$ with the additional rules and preference statements
described in this section,
$K \models_{\tip min} Q$ if and only if $\pi_Q$ is in all the optimal solutions computed by {\em asprin}
for $\Pi_{\tip min}(K,Q)$.

Observe that deciding the existence of a $\tip$-minimal answer set of $\Pi(K)$ falsifying $\pi_Q$ is a problem in $\Sigma^P_2$ (see Appendix E, Proposition 6)
and it could also be solved by direct encoding in Disjunctive Datalog with negation \cite{Eiter97} under the stable model
semantics.
By Proposition 5 in Appendix E, checking whether $K \models_{\tip min} Q$ is then in $\Pi^P_2$, and, given the hardness result in Theorem \ref{th:lowerBound}, it is $\Pi^P_2$-complete.

In a similar way, answer set preferences in the {\em asprin} framework allow to capture ABox minimization,
i.e. minimization of the ranks of named individuals  (assigning an higher priority to concept rank minimization).
In particular, this can be done introducing a statement:
\begin{tabbing}
$ \mathit{\#preference(p_{a_i},less(weight)) \{  X,X :: rank(a_i,X) : possrank(X) \} } $
\end{tabbing}
for each  $a_i \in N_{I,K}$, that defines a preference, named $\mathit{p_{a_i}}$,
for a smaller rank of $\mathit{a}$;
and replacing $ \mathit{ \#optimize(p\mbox{-}tbox) } $ with the statements:
\begin{tabbing}
$ \mathit{ \#preference(p\mbox{-}abox,pareto) \{ name(p_{a_1}) ; \ldots ; name(p_{a_n}) \} } $ \\
$ \mathit{ \#preference(p\mbox{-}lex,lexico) \{ 2 :: name(p\mbox{-}tbox) ; 1 :: name(p\mbox{-}abox) \} } $ \\
$ \mathit{ \#optimize(p\mbox{-}lex) } $
\end{tabbing}
which require an optimal solution with respect to the lexicographic combination $p\mbox{-}lex$ of
the {\em pareto} combination $p\mbox{-}tbox$ of the minimization of concept ranks, and,
with smaller priority,
the {\em pareto} combination $p\mbox{-}abox$ of the minimization of individual ranks.

In Table \ref{runningtimes} we report some results about the actual execution of the framework in {\em asprin}.
We use Example 1 as a basis, using also minimization of the rank of individuals, as described above.
We report the running times (in seconds) for variants of the example as the $\mathit{ABox}$ grows, replicating (up to 8 times) the $\mathit{ABox}$ of Example 1,
i.e., adding $\mathit{Student}(\mathit{mary'}), ~
\mathit{friendOf}(\mathit{mario'}, \mathit{mary'})$, and so on;
and running times for variants where the whole $\mathit{KB}$ grows, replicating, again up to 8 times, the entire example $\mathit{KB}$,
i.e., adding
$\tip(\mathit{Italian'}) \sqsubseteq \mathit{ \exists hasHair'.\{Black'\}}, \ldots$ as well as
$\mathit{Student'}(\mathit{mary'})$, and so on.

It can be seen that the basic example requires a small but non-negligible running time (0.82 seconds); the approach scales up well (first row) with respect to the $\mathit{ABox}$, and not equally well in case (second row) both
the $\mathit{ABox}$ and $\mathit{TBox}$ grow.

\begin{table}[t]
\begin{tabular}{ c || c c  c  c  c }
\hline
\hline
   \ \  & 1x &    2x     &  4x     &  6x     &  8x     \\
\hline
\hline
Replication of $\mathit{ABox}$ ~ ~ ~
 & 0.82
 & 1.01
 & 1.34
 & 1.63
 & 1.90
\\
\hline
\hline
Replication of $\mathit{KB}$ ~ ~ ~
 & 0.82
 & 1.96
 & 3.87
 & 27.28
 & 40.62
 \\
\hline
\hline
\end{tabular}
\caption{Some scalability results for Example 1}
\label{runningtimes}
\end{table}

\section{Conclusions and Related Work}

In this paper we have shown that Answer Set Programming can be used for reasoning under a minimal model semantics
in a rational extension 
of the low complexity description logic $\sroel$, which underlies the OWL EL ontology language.
In particular, we have defined an ASP encoding $\Pi(K)$ of a knowledge base $K$ so that the answer sets of $\Pi(K)$
correspond to small (finite and polynomial)  models of $K$.
The encoding is based on the materialization calculus for instance checking in Datalog by Kr\"{o}tzsch \shortcite{KrotzschJelia2010}
for the logic $\sroel$.
We propose a $\tip$-minimal model semantics which is an alternative to the minimal canonical model semantics in \cite{AIJ15},
but which coincides with it when minimal canonical models of the KB exist
and their ranking of concepts agrees with the ranking computed by rational closure.
The advantage of the $\tip$-minimal model semantics is that it can be defined also for some KBs for which no minimal canonical model exists.
We show that instance checking under $\tip$-minimal entailment in $\sroel^{\Ra \ }\tip$ is \textsc{$\Pi^P_2$}-complete
and we use the {\em asprin} framework \cite{BrewkaIJCAI15} for Answer Set Preferences to compute minimal entailment.
The approach is extended to deal with ABox minimization,
by minimizing the ranks of individual names, and can be used to experiment alternative notions of minimization.

Tableaux-based proof methods for a preferential extension of low complexity DLs including $\elbot$
have been studied in \cite{GiordanoLPNMR09}, based on interpretations that are not required to be modular,
and on minimizing $\neg \Box \neg C$ concepts.
For such a logic, in \cite{ijcai2011}  it is shown that minimal entailment
is \textsc{ExpTime}-hard already for simple KBs,
similarly to circumscriptive KBs \cite{Bonatti2011}.

Nonmonotonic extensions of DLs  include the formalisms for combining DLs with logic programming rules,
such as for instance, \cite{Eiter2008}, \cite{rosatiacm}, \cite{KnorrECAI12} and Datalog +/- \cite{Gottlob14}.
In \cite{bonattiAIJ15} a non monotonic extension of DLs is proposed based on a notion of overriding and supporting normality concepts.
In particular, it preserves the tractability of low complexity DLs, including ${\el}^{++}$ and $DL$-$lite$.
In \cite{KnorrECAI12} a general DL language is introduced,
which extends ${\sroiq}$ with nominal schemas and epistemic operators as defined in \cite{rosatiacm},
and encompasses some of the most prominent nonmonotonic rule languages, including ASP.
The CKR framework  \cite{Bozzato14}, 
based on {\em SROIQ-RL},
allows for defeasible axioms with local exceptions. 
It is shown that instance checking over a CKR reduces to (cautious) inference under the answer sets semantics.

The work in this paper could provide a starting point for devising more effective approaches for computing $\tip$-minimal  entailment or alternative notions of defeasible entailment in low complexity DLs.
In particular, for the fragment of $\sroelrt$ for which $\tip$-minimal entailment provides a characterization of
the rational closure of the KB, that we expect to be larger than the intersection with $\alcrt$,
computing  $\tip$-minimal entailment can be made more efficient through the rational closure construction,
since rational entailment is polynomial \cite{DL2016+CILC}.
To this purpose, a combination with the polynomial Datalog encoding of entailment in $\sroelrt$ in \cite{DL2016+CILC}
can be exploited.
Future work may also include 
optimizations based on modularity as in \cite{BonattiSWC15}, 
as well as considering
refinements of the rational closure, such as the lexicographic closure, introduced by Lehmann \shortcite{Lehmann95}
and extended to $\alc$ in \cite{Casinistraccia2012}, and the relevant closure proposed in \cite{Casini14}.
The combination of low complexity DLs and rule languages can provide a convenient setting in which alternative approaches to the definition of exceptions in DLs can be compared, and can as well be a source of challenging problems for ASP solvers.

\smallskip
{\bf Acknowledgement}.
This research has been partially supported by INDAM - GNCS Project 2016 {\em Ragionamento Defeasible nelle Logiche Descrittive}.

\bibliography{gmt,biblioKR16}

\begin{appendix}
\section{Proofs of Theorems 1 and 2} \label{appendix A}

{\em Theorem 1 }
{\em
(Small Rank)

\noindent
Let $K$=(TBox,RBox,ABox) be a normalized $\sroelrt$ knowledge base. Given any
model $\mathcal{M}=(\Delta, <, \cdot^I )$ of $K$, 
there exists a model
$\mathcal{\emme'}=(\Delta, <', \cdot^{I'})$ of $K$ (over the extended language) such that, for all $x \in \Delta'$:
(i) $k_{\emme'}(x) \leq max_K$;
(ii) for all $C \in N_C$, $x \in C^{I'}$ iff $x \in C^{I}$; and
(iii) for all $C \in C_{K}$, $x \in (\tip(C))^{I'}$ iff $x \in (\tip(C))^{I}$.
}

\begin{proof}
We define the model $\emme'$ over the domain $\Delta$ by letting $\cdot^{I'}=\cdot^{I}$,
while changing the rank of the elements in $\Delta$.
What is preserved from $\emme$ is the relative order of the ranks of the typical $C$ elements, for $C \in C_{K}$.
Remember that, from the definition of the rank of a concept in a model, 
$k_\emme(C)$  is equal to the rank of all the typical $C$'s in $\emme$
(which must have all the same rank).
Let us partition the set $C_{K}$ according to the ranks of the concepts in $\emme$:

\noindent
$H_0= \{$ $C \in C_{K} \mid$ there is no $D \in C_{K}$ with
$k_ {\emme}(D)< k_\emme(C)$\}

\noindent
$H_i= \{$ $C \in C_{K}-(H_0\cup \ldots \cup H_{i-1}) \mid$ there is no
$D \in C_{K}-(H_0\cup \ldots \cup H_{i-1})$ with  $k_ {\emme}(D)< k_ \emme(C)$\}

\noindent
As the set $C_{K}$ is finite and its cardinality is $max_K$,
there is some minimum $n < max_K$, such that 
$H_{n+1} = \emptyset$.

We define the relation $<'$ by setting the rank of all the domain elements in $\emme'$  between $0$ and $n+1$.
In particular, we  want to let the rank of all the typical $C$ elements to be $i$, if $C \in H_i$.
For all $x \in \Delta$:\\
- if $k_\emme(x) \leq k_ \emme(C)$ for some $C \in H_0$, then let $k_{\emme'}(x)= 0$;\\
- if 
 $k_ \emme(B) < k_\emme(x) \leq k_ \emme(C)$
for some $B \in H_{i-1}$ and $C \in H_{i}$  ($0< i \leq n$),
then let $k_{\emme'}(x)= i$; \\
- if 
 $k_ \emme(B) < k_\emme(x)$ for some $B \in H_n$,
then let $k_{\emme'}(x)= n+1$.

In particular, we let the rank of all the typical $C$ elements to be $i$, if $C \in H_i$.
In fact, if $x \in (\tip(C))^I$ then $k_{\emme}(x)= k_ \emme(C)$.  In case  $C \in H_{i}$,
then $k_{\emme'}(x)= i$.

Changing the ranks as above cannot make a domain element, which is a typical $C$ (for some $C \in C_{K}$),
become a nontypical $C$ element.
In fact, if $x \in (\tip(C))^I$, then for all $y$ such that $k_\emme(y) <k_\emme(x)$, $y \not \in C$.
Suppose a typical $C$ element $x$ gets the rank $i$ in $\emme'$ (as $C \in H_i$).
Some $y$ can get in $\emme'$ the same rank as $x$ if
$k_\emme(B) < k_\emme(y) \leq k_\emme(C)$,
for some $B \in H_{i-1}$. However, even if the rank of $y$ becomes $i$,
$x$ remains a typical $C$ element.
Also, it is not the case that a nontypical $C$ element $z$ (for $C \in C_K$) can become  a typical $C$ element.
In fact, one such $z$ must have a rank $k_\emme(z)$ greater than the rank of any typical $C$ element $x$,
i.e., $k_\emme(x) < k_\emme(z)$.
If  $x$ gets rank $i$ in $\emme'$, since $C \in H_i$, then (by definition of $\emme'$) $z$ gets a rank higher then $i$.
Of course, this is not true for the concepts $C \not \in C_{K}$. However, we can include as well in the set $C_{K}$
all the concepts $C$ such that $\tip(C)$ might occur in a query.
\end{proof}


\noindent
{\em Theorem 2}
{\em
Let $K$ be a knowledge base satisfying the following conditions:

(i) a canonical model of $K$ exists;

(ii) the ranking $K_{\emme}$ of each canonical model $\emme$ of $K$ is the same as the one determined

$\mbox{\ \ \ \ }$ by the Rational Closure construction;

\noindent
and let $Q$ be an inclusion $\tip(C) \sqsubseteq D$
(where $C$ and $D$ are non-extended concepts).
Then, $K \models_{\tip min} \tip(C) \sqsubseteq D$ iff $K \models_{min} \tip(C) \sqsubseteq D$.
}

\begin{proof}
{\em (If)} By contraposition. Suppose that $K \not \models_{\tip min} Q$, i.e.
there is a $\tip$-minimal model $\emme$ of $K$ which falsifies $Q$.
Let us consider any minimal canonical model $\emme'$ of $K$ (there is one by (i)).
$\emme'$ must give the same ranks as $\emme$ to the concepts $C \in {\cal T}_{K,Q}$.
First it is not the case that $\emme' \prec_{\tip} \emme$,
otherwise $\emme$ would not be a $\tip$-minimal model.
Also, it is not the case that
there is a concept $C \in {\cal T}_{K,Q}$ such that
$k_{\emme}(C) <  k_{\emme'}(C)=rank(C)$, as
the rank of a concept in any model of $K$
cannot be lower than $rank(C)$, the rank of $C$ in the Rational Closure\footnote{Observe that,
the rank of a concept $C$ can be determined in the
rational closure construction for a KB in $\sroelrt$,
by iteratively verifying exceptionality of the concept $C$ with respect to a set of inclusions $E_i$ according to the
iterative construction in \cite{AIJ15}: {\em $C$ is exceptional wrt. $E_i$} iff $E_i \models_{sroelrt} \tip(\top) \sqcap C \sqsubseteq \bot$.
For a concept $C \wedge \neg D$, where $C$ and $D$ are non extended concepts,
{\em $C \wedge \neg D$ is exceptional wrt. $E_i$} iff $E_i \models_{sroelrt} \tip(\top) \sqcap C \sqsubseteq D$.}
(this property holds for $\sroelrt$ as it holds for $\alcrt$ \cite{AIJ15} and for $\shiqrt$ \cite{DL2014}).
If there is a concept $C \in {\cal T}_{K,Q}$ such that
$k_{\emme'}(C) <  k_{\emme}(C)=rank(C)$, then
as we have excluded that $\emme' \prec_{\tip} \emme$,
there must be a concept $C' \in {\cal T}_{K,Q}$ such that
$k_{\emme}(C') <  k_{\emme'}(C')=rank(C')$,
(i.e., the two models $\emme$ and $\emme'$ must be incomparable wrt. $\prec_{\tip}$).
But we have already seen that it not possible that the rank of $C'$ in a model is lower
than the rank of $C'$ in the rational closure.
Thus, the minimal canonical model $\emme'$ assigns to the concepts in ${\cal T}_{K,Q}$
the same rank as $\emme$.

We have to show that $\emme'$ falsifies the query $Q$.
Let $Q$ be $\tip(C) \sqsubseteq D$.
As $\emme$ falsifies $\tip(C) \sqsubseteq D$, there is 
an element $x \in \Delta$ such that  $x \in (T(C))^I$ ($x$ is a typical $C$ element in $\emme$) and $x \not \in D^I$.
Hence, $x \in (C \sqcap \neg D)^I$. Let $k_{\emme}(x)=i$ (and hence $k_{\emme}(C)=i$).
As $\emme'$ is a canonical model, $\emme'$ must contain a domain element $y \in (C \sqcap \neg D)^{I'}$.
Clearly, $k_{\emme'}(C \wedge \neg D) \geq k_{\emme'}(C)$.
If $k_{\emme'}(C \wedge \neg D)=i$, then $y \in \tip(C)^{I'}$
(as $C$ has the same rank $i$ in $\emme$ and in $\emme'$), and $\emme'$ falsifies $Q$.
We show that assuming that $k_{\emme'}(C \wedge \neg D)=j>i$,
leads to a contradiction.
By hypothesis (ii) $\emme'$ assigns to concepts the same rank as the rational closure,
hence $rank(C \wedge \neg D)=j>i$ in the rational closure.
This contradicts the fact that $k_{\emme}(C \wedge \neg D)=i$,
as the rank of a concept in a model of $K$
cannot be lower than the rank of that concept in the Rational Closure.

{\em (Only If)} By contraposition.
Let $\emme$ is a minimal canonical model of $K$ falsifying $Q$. We want to show that
there is a $\tip$-minimal model $\emme'$ falsifying $Q$.
We can show that $\emme$ is itself a $\tip$-minimal model of $K$ (falsifying $Q$).
Clearly, $\emme$ is a $\tip$-complete model of $K$.
If $\emme$ were non-minimal wrt.   $\prec_{\tip}$,
there would be a model $\emme' \prec_{\tip} \emme$.
In this case, there would be a $C \in {\cal T}_{K,Q}$ such that $k_{\emme'}(C) <  k_{\emme}(C)$.
This is not possible, due to the property that the rank of a concept $C$ in a model of $K$
cannot be lower than $rank(C)$, the rank of the concept $C$ in the Rational Closure.
As, from hypothesis (ii), $k_{\emme}(C)=rank(C)$, it is not the case that $k_{\emme'}(C) <  k_{\emme}(C)$.
\end{proof}


\section{Proof of Theorem 3: Lower Bound for $\tip$-minimal entailment} \label{Appendix Lower Bound}

In this section we show that  the
problem of deciding instance checking under the $\tip$-minimal model semantics is a
\textsc{$\Pi^P_2$}-hard problem for $\sroelrt$ knowledge bases.
To show this, we provide a reduction of the minimal entailment problem of
{\em positive disjunctive logic programs}, which has been proved to be a
 \textsc{$\Pi^P_2$}-hard problem by Eiter and Gottlob in \cite{Eiter95}.
A similar reduction has been used
to prove  \textsc{$\Pi^P_2$}-hardness of entailment for Circumscribed
Left Local $\mathit{EL}^\bot$ knowledge bases in \cite{Bonatti2011}.

Let $PV = \{p_1, \ldots, p_n \}$ be a set of propositional variables.
A clause is formula $l_1 \vee \ldots \vee l_h$, where each literal $l_j$  is either a
propositional variable $p_i$ or its negation $\neg p_i$.
A positive disjunctive logic program (PDLP) is a set of clauses $S= \{\gamma_1, \ldots, \gamma_m\}$,
where each $\gamma_j$ contains at least one positive literal.
A truth valuation for $S$ is a set $I \subseteq PV$,
containing the propositional variables which are true. A truth valuation is a model of $S$
if it satisfies all clauses in $S$.
For a literal $l$, we write $S \models_{min} l$ if and only if every minimal model (with respect to subset inclusion)
of $S$ satisfies $l$. The minimal-entailment problem can be then defined as follows:
given a PDLP $S$ and a literal $l$, determine whether  $S \models_{min} l$.
In the following we sketch the reduction of the minimal-entailment problem for a  PDLP $S$
to the instance checking problem under $\tip$-minimal entailment, from a knowledge base $K$ constructed from $S$.

We define a KB $\mathit{K = (TBox,RBox,ABox)}$
in $\sroelrt$ as follows.
We introduce a concept name $P_h\in N_C$ for each variable $p_h\in PV$ ($h=1,\ldots,n$).
Also, we introduce in $N_C$ an auxiliary concept $H$, a concept name $D_S$ associated with the set of clauses $S$,
and a concept name $D_j$ associated with each clause $\gamma_j$ in $S$ ($j=1,\ldots,m$).
We let $a \in N_I$ be an individual name,
and we define $K$ as follows:

$\mathit{RBox}=\emptyset$,

$\mathit{ABox}=\{ P_h (a), h=1,\ldots,n \} \cup \{\tip(H)(a), D_S(a)\}$,

\noindent
and $\mathit{TBox}$ contains the following inclusions
(where $ C_i^j$ and $\overline{ C_i^j}$ are concepts associated with each literal $l_i^j$ occurring in $\gamma_j = l_1^j \vee \ldots \vee l_k^j$, as defined below):

(1) $\tip(\top) \sqcap H \sqsubseteq \bot$

(2) $\{a\} \sqcap C_i^j \sqsubseteq D_j$ \ \ \ for all $\gamma_j = l_1^j \vee \ldots \vee l_k^j$ in $S$

(3) $\{a\} \sqcap D_j \sqcap \overline{ C_1^j}  \sqcap \ldots \sqcap \overline{ C_k^j} \sqsubseteq \bot$
\ \ \ for all $\gamma_j = l_1^j \vee \ldots \vee l_k^j$ in $S$

(4) $\{a\} \sqcap D_1  \sqcap \ldots  \sqcap D_m \sqsubseteq D_S$  

(5) $\{a\} \sqcap D_S \sqsubseteq  D_1  \sqcap \ldots  \sqcap D_m$  

\noindent
for each $h=1,\ldots,n$,
 $j=1,\ldots,m$,
and where $C_i^j$ is defined as follows:
\[
   C_i^j=
\begin{cases}
	\tip(P_h) \mbox{\ \ \ \ \ \ \ \ \ \ \ \ \ \ \ \ \ \ \ \ \ \ \ \ \ \  if } l_i^j=p_h\\
    	\exists U.(\tip(\top) \sqcap P_h) \mbox{\ \ \ \ \ \ \ \ \ if } l_i^j=\neg p_h
\end{cases}
\]

\[
   \overline{C_i^j}=
\begin{cases}
	\exists U.(\tip(\top) \sqcap P_h) \mbox{\ \ \ \ \ \ \ \ \ if } l_i^j=p_h\\
    	\tip(P_h) \mbox{\ \ \ \ \ \ \ \ \ \ \ \ \ \ \ \ \ \ \ \ \ \ \ \ \ \  if }  l_i^j=\neg p_h
\end{cases}
\]
where $U$ is the universal role.
\noindent
Let us consider any model $\emme$$=\sx\Delta, <, \cdot^I\dx$ of $K$.
Observe that, all the $\tip(\top)$ elements are all $\neg H$ elements.
Hence, $a^I$ (being a typical $H$) must have rank greater then 0,
and it will have rank 1 in all $\tip$-minimal models.
The $\tip$-minimal models of $K$ satisfying $D_S(a)$ are intended to correspond
to the (propositional) minimal interpretations $J$ satisfying $S$.
Roughly speaking, the concepts $P_h$ such that
$a^I \in (\tip(P_h))^I$ in $\emme$ correspond to the variables $p_h$ in the minimal interpretation $J$ satisfying $S$.
In any $\tip$-minimal model of $K$, 
either $P_h$ has rank 0 (and $a$ is not a typical $P_h$),
or $P_h$ has rank 1 (and $a$ is a typical $P_h$).
Clearly, by $\tip$-minimality, a model of $K$ 
in which the ranking of a set of $P_h$'s is 0,
is preferred to the models in which the ranking of some of those $P_h$'s is higher (i.e. 1).
This captures the subset inclusion minimality in the interpretations of the positive disjunctive logic program $S$.
Inclusions (2)-(5) bind the truth values of the $P_h(a)$ to the truth values of the clauses in $S$ and of their conjunction. 
The assertion $D_S(a)$ in $ABox$ is required to select only those interpretations satisfying the set $S$ of disjunctions.
Observe also that any $\tip$-minimal model must contain al least a $P_h$ element, for each $h=1,\ldots,n$,
as $P_h$ is a consistent concept.

In  any minimal canonical model $\emme$ of $K$ satisfying $D_S(a)$:
either $a^I \in (\tip(P_h))^I $ or \linebreak $\mathit{a^I \in  (\exists U.(\tip(\top) \sqcap \tip(P_h)))^I}$.
Hence, for $a^I$ the two concepts in the definition of $C_i^j$ are disjoint and complementary,
and $\overline{C_i^j}$ is actually the concept representing the complement of $C_i^j$.
Given a set $S$ of clauses and a literal $L$, the following holds:

\medskip

\noindent
{\em Proposition 4 }

\noindent
Given a set $S$ of clauses and a literal $L$,
\begin{center}
$S \models_{min}  L$ \ \ \  if and only if \ \ \ \ $K \models_{\tip min} C_L(a)$
\end{center}
where $C_L$ is the concept associated with $L$, i.e., $C_L=\tip(p_h)$ if $L=p_h$, and
$C_L=\exists U.(\tip(\top) \sqcap P_h)$ if $L=\neg p_h$.

\medskip

From the reduction above and the fact that minimal entailment for PDLP is
 \textsc{$\Pi^P_2$}-hard \cite{Eiter95}, it follows that
 minimal entailment under $\tip$-minimal model semantics is $\Pi^P_2$-hard,
 i.e. Theorem 3 holds.


\section{Calculus for instance checking in $\sroel$ }
\label{Appendix_Calculus}

We report the calculus for $\sroel$ instance checking from \cite{KrotzschJelia2010} used in section 5 and, with a small variant, in section 4.
The representation of a knowledge base ({\em input translation}) is as follows, where, to keep a DL-like notation, we do not follow the ASP convention where variable names start with uppercase;
in particular, $A$, $B$ $C$, and $R$, $S$, $T$, are intended as ASP constants corresponding to the same class/role names in $K$:

\vspace{-0.3cm}
\begin{tabbing}
$ \mathit{ \exists R . Self \sqsubseteq C }$ \= \kill  \\
\> $\mathit{ a \in N_I }$ \'             $\mapsto \mathit{ nom(a) } $ \\
\> $\mathit{ C \in N_C }$ \'             $\mapsto \mathit{ cls(C) } $ \\
\> $\mathit{ R \in N_R }$ \'             $\mapsto \mathit{ rol(R) } $ \\
\> $\mathit{C(a)}$ \'                    $\mapsto \mathit{ subClass(a,C) } $ \\
\> $ \mathit{ R(a,b)}$ \'                $\mapsto \mathit{supEx(a,R,b,b) } $ \\
\> $ \mathit{ \top \sqsubseteq C}$ \'    $\mapsto \mathit{top(C)} $ \\
\> $ \mathit{ A \sqsubseteq \bot}$ \'    $\mapsto \mathit{bot(A) } $ \\
\> $ \mathit{ \{a\} \sqsubseteq C  }$ \' $\mapsto \mathit{subClass(a,C) } $ \\
\> $ \mathit{ A \sqsubseteq \{c\}  }$ \' $\mapsto \mathit{subClass(A,c) } $ \\
\> $ \mathit{ A \sqsubseteq C  }$ \'     $\mapsto \mathit{subClass(A,C) } $ \\
\> $ \mathit{ A \sqcap B \sqsubseteq C }$ \'            $\mapsto \mathit{subConj(A,B,C) } $ \\
\> $ \mathit{ \exists R . Self \sqsubseteq C  }$ \'     $\mapsto \mathit{subSelf(R,C)  } $ \\
\> $ \mathit{ A \sqsubseteq \exists R . Self  }$ \'     $\mapsto \mathit{supSelf(A,R)  } $ \\
\> $ \mathit{ \exists R . A \sqsubseteq C  }$ \'        $\mapsto \mathit{subEx(R,A,C)  } $ \\
\> $ \mathit{ A \sqsubseteq \exists R . B   }$ \'    $\mapsto \mathit{supEx(A,R,B,aux_i)  }$ \\
\> $ \mathit{ R \sqsubseteq T  }$ \'    $\mapsto \mathit{ subRole(R,T)  } $ \\
\> $ \mathit{ R \circ S  \sqsubseteq T  }$ \'    $\mapsto \mathit{ subRChain(R,S,T)  } $ \\
\> $ \mathit{ R \sqsubseteq C \times D  }$ \'    $\mapsto \mathit{ supProd(R,C,D)  } $ \\
\> $ \mathit{ A \times B \sqsubseteq R  }$ \'    $\mapsto \mathit{ subProd(A,B,R) } $ \\
\> $ \mathit{ R \sqcap S \sqsubseteq T  }$ \'    $\mapsto \mathit{ subRConj(R,S,T) } $ \\
\end{tabbing}
\vspace{-0.3cm}
In the translation of $ \mathit{ A \sqsubseteq \exists R . B   }$,
$\mathit{aux_i}$  is a new constant, different for each axiom of this form.

The {\em inference rules} (included in $\Pi_{IR}$ in section 4) are the following\footnote{Here, $u, v, x, y, z, w$, possibly with suffixes, are ASP variables.}:
\vspace{-0.3cm}
\begin{tabbing}
$(10)$ \= \kill \\
$(1) ~ \mathit{inst(x, x) \leftarrow nom(x) }  $\\
$(2) ~ \mathit{self(x, v) \leftarrow  nom(x), triple(x, v, x) } $\\
$(3) ~ \mathit{inst(x, z) \leftarrow top(z), inst(x, z') } $\\
$(4) ~ \mathit{\bot \leftarrow bot(z), inst(u, z) } $\\
$(5) ~ \mathit{inst(x, z) \leftarrow subClass(y, z), inst(x, y) }  $\\
$(6) ~ \mathit{inst(x, z) \leftarrow subConj(y1, y2, z), inst(x, y1), inst(x, y2) } $\\
$(7) ~ \mathit{inst(x, z) \leftarrow subEx(v, y, z), triple(x, v, x'), inst(x', y) } $\\
$(8) ~ \mathit{inst(x, z) \leftarrow subEx(v, y, z), self(x, v), inst(x, y) } $\\
$(9) ~ \mathit{triple(x, v, x') \leftarrow supEx(y, v, z, x'), inst(x, y) } $\\
$(10) ~ \mathit{inst(x', z) \leftarrow supEx(y, v, z, x'), inst(x, y) } $\\
$(11) ~ \mathit{inst(x, z) \leftarrow subSelf(v, z), self(x, v) } $\\
$(12) ~ \mathit{self(x, v) \leftarrow supSelf(y, v), inst(x, y) } $ \\
$(13) ~ \mathit{triple(x, w, x') \leftarrow subRole(v, w), triple(x, v, x') } $ \\
$(14) ~ \mathit{self(x,w) \leftarrow subRole(v, w), self(x, v) }  $ \\
$(15) ~ \mathit{triple(x, w, x'') \leftarrow subRChain(u, v,w), triple(x, u, x'), triple(x', v, x'') } $ \\
$(16) ~ \mathit{triple(x, w, x') \leftarrow subRChain(u, v,w), self(x, u), triple(x, v, x') }  $ \\
$(17) ~ \mathit{triple(x, w, x') \leftarrow subRChain(u, v, w), triple(x, u, x'), self(x', v) }  $ \\
$(18) ~ \mathit{triple(x, w, x) \leftarrow subRChain(u, v,w), self(x, u), self(x, v) }  $ \\
$(19) ~ \mathit{triple(x, w, x') \leftarrow subRConj(v1, v2,w),  triple(x, v1, x'), triple(x, v2, x') }  $ \\
$(20) ~ \mathit{self(x,w) \leftarrow  subRConj(v1, v2,w), self(x, v1), self(x, v2) } $ \\
$(21) ~ \mathit{triple(x, w, x') \leftarrow subProd(y1, y2,w), inst(x, y1), inst(x', y2) }  $ \\
$(22) ~ \mathit{self(x,w) \leftarrow  subProd(y1, y2,w), inst(x, y1), inst(x, y2) } $ \\
$(23) ~ \mathit{inst(x, z1) \leftarrow supProd(v, z1, z2), triple(x, v, x') } $ \\
$(24) ~ \mathit{inst(x, z1) \leftarrow supProd(v, z1, z2), self(x, v) } $ \\
$(25) ~ \mathit{inst(x', z2) \leftarrow  supProd(v, z1, z2), triple(x, v, x') } $ \\
$(26) ~ \mathit{inst(x, z2) \leftarrow  supProd(v, z1, z2), self(x, v) } $ \\
$(27) ~ \mathit{inst(y, z) \leftarrow  inst(x, y), nom(y), inst(x, z) } $ \\
$(28) ~ \mathit{inst(x, z) \leftarrow  inst(x, y), nom(y), inst(y, z) } $ \\
$(29) ~ \mathit{triple(z, u, y) \leftarrow inst(x, y), nom(y), triple(z, u, x) } $
\end{tabbing}

\noindent
The version of the calculus in \cite{KrotzschJelia2010}, used in Section 5, contains the rule:

\noindent
$(4b) ~ \mathit{inst(x, y) \leftarrow bot(z), inst(u, z), inst(x, z'), cls(y) } $

\noindent
instead of rule (4) above.

\section{Proofs for Section 4} \label{appendix:Section4}

\subsection{Proof of Proposition 2} \label{appendix:Prop2}

{\em Proposition 2.}
{\em
Given a normalized knowledge base $K$ and a query $Q$, if there is an answer set $S$ of the ASP program $\Pi(K) \cup \{- \pi_Q\}$,
then there is a model $\emme=(\Delta, <, \cdot^I)$ of $K$  such that  $Q$ is  not satisfied in $\emme$.
}

\medskip
\noindent
The proof is similar to the one for Lemma 3  in \cite{jeliaReport}, which proves the completeness of
the materialization calculus for $\sroel$ by contraposition, building a model of the KB from the minimal Herbrand model of the Datalog encoding.
Here, given the answer set $S$ of the program $\Pi(K) \cup \{- \pi_Q\}$ we build the model $\emme$ falsifying $Q$ exploiting the information in $S$.

%
%
%
In particular, we construct the domain of $\emme$ from the set $Const$ including all the name constants $c \in N_I$
as well as all the auxiliary constants occurring in the ASP program $\Pi(KB,Q)$,  
defining an equivalence relation over constants and using equivalence classes
to define domain elements. For readability, we write $\auxARC$ and $aux_C$, respectively, for the constants
associated with inclusions $A \sqsubseteq \exists R.C$ and with the typicality concepts $\tip(C)$.
Observe that the answer set $S$ contains all the details about the definition of the ranking of the domain elements
that can be used to build the model $\emme$.

First, let us define a relation $\approx$ between the constants in $Const$:

\medskip

\noindent
{\em Definition 7 }

\noindent
Let $\approx$ be the reflexive, symmetric and transitive closure of the relation
$\{ (c,d) \mid inst(c,d) \in S$, for $c \in Const$ and  $d \in N_I\}$.

\medskip

\noindent
It can be proved that:
\begin{lemma} \label{lemma:approx}
Given a constant $c$ such that $c \approx a$ for $a \in N_I$,
if  $inst(c,A)$ ($triple(c,R,d)$, $triple(d,R,c)$, $self(c,R)$, $rank(c,k)$) is in $S$,
then $inst(a,A)$ ($triple(a,R,d)$, $triple$ $(d,R,a)$, $self(a,R)$, $rank(a,k)$) is in $S$.
\end{lemma}
The proof is similar to the proof of Lemma 2 in  \cite{jeliaReport}.
For the predicate {\em rank}, the proof exploits rule (46).
The vice-versa of Lemma  \ref{lemma:approx} only holds for some of the predicates, namely:
\begin{lemma} \label{lemma:approx2}
Given a constant $c$ such that $c \approx a$ for $a \in N_I$,
if  $inst(a,A)$ ($triple(a,R,d)$, $rank(a,k)$) is in $S$,
then $inst(c,A)$ ($triple(c,R,d)$,  $rank(c,k)$) is in $S$.
\end{lemma}

Now, let $[c]=\{ d \mid d \approx c \}$ denote the equivalence class of $c$;
we define the domain $\Delta$ of the interpretation $\emme$ as follows:
 $\Delta=  \{[c] \mid c \in N_I\} \cup \{\wARC_1, \wARC_2 \mid$ $inst(\auxARC,e) \in S$ for some $e$ and there is no $d \in N_I$ such that $\auxARC \approx d\}$
 $ \cup \{z_C^1 , z_C^2 \mid$ $inst( aux_C,e) \in S$ for some $e$ and there is no $d \in N_I$ such that $aux_C \approx d\}$.
Two copies of auxiliary constants are introduced, as in \cite{jeliaReport}, to handle $\mathit{Self}$ statements.

For each element $e \in \Delta$, we define a projection $\prj(e)$ to $Const$ as follows:

- $\prj([c])=c$;

- $\prj(\wARC_i)=\auxARC$, i=1,2;

- $\prj(z_C^i)=aux_C, i=1,2;$

\noindent
We define the interpretation of individual constants, concepts and roles over $\Delta$ as follows:\\
$\mbox{\ \ }$ - for all $c \in N_I$, \ $c^I =[c]$;\\
$\mbox{\ \ }$ - for all $d \in \Delta$, \   $d \in A^{I}$ iff $\mathit {inst}(\prj(d), A) \in S$; \\
$\mbox{\ \ }$ - for all $d, e \in \Delta$, \ $(d,e) \in R^{I}$ iff ($\mathit{triple}(\prj(d), R, \prj(e)) \in S$ and  $d \neq e$)\\
$\mbox{\ \ \ \ \ \ \ \ \ \ \ \ \ \ \ \ \ \ \ \ \ \ \ \ \ \ \ \ \ \ \ \ \ \ \ \ \ \ \ \ \ \ \ \ \ \ \ \ \ \ \ \ \ \ \ \ }$  or ($\mathit{self}(\prj(d), R) \in S$ and  $d = e$).

We define the rank of the domain elements in $\Delta$ in agreement with the extension of the $rank$ predicate in $S$:\\
$\mbox{\ \ }$ - for all $d \in \Delta$, \ \ $k_{\emme}(d) = h$,  iff $\mathit{rank}(\prj(d), h) \in S$.

\noindent
In particular, $z_C$ has rank $h$ if $\mathit{rank}(aux_C, h) \in S$ and
$\wARC$ has rank $h$ if $\mathit{rank}$ $(\auxARC, h) \in S$.
The rank function $k_{\emme}([c])$ is well defined.
In fact, there is exactly one $h$ such that $\mathit{rank}(\prj(d),h)\in S$ for each $\prj(d)$ (rules (36) and (37)).
It is easy to see by Lemma \ref{lemma:approx}  and Lemma \ref{lemma:approx2} that, when $aux_C \approx a$ ($a \in N_I$), i.e., $auc_C \in [a]$,
we have $k_{\emme}([a])=h$ iff $\mathit{rank}(aux_C, h) \in S$.
As a consequence, all the concepts $C$ such that $\tip(C)$ occurs in $K$ (or in $Q$) have that same rank in $\emme$ and in $S$.



\medskip
To conclude the proof of Proposition 2 it suffices to prove that $\emme$ is a model of KB, i.e. it satisfies all the
axioms in KB.
The proof is as in  \cite{jeliaReport} (see Lemma 2),
except that we have to consider the additional axioms $A \sqsubseteq \tip(B)$ and $ \tip(B) \sqsubseteq C$.

For $A \sqsubseteq \tip(B)$ in KB, we have $\mathit{supTyp(A,B)} \in S$.
Let us assume that $d \in A^I$. We want to prove that $d \in (\tip(B))^I$.
By construction  $\mathit {inst}(\prj(d), A) \in S$.
By rule (30), $\mathit {typ}(\prj(d), B) \in S$.
By rule (47), $\mathit {inst}(\prj(d), B) \in S$, i.e., $d \in B^I$.
Let $\mathit{rank(\prj(d),h) }\in S$, i.e. $k_\emme(d)=h$.

To show that $d$ is a typical $B$,
we have to show that, for all the domain elements $e$ with rank $j<h$,
$e \not \in B^I$.
Given that $\mathit {typ}(\prj(d), B)$ and $\mathit{rank(\prj(d),h) }$ are in $S$, from rule (49),
$\mathit {box\_neg}(h, B) \in S$. From the repeated application of rule (41), $\mathit {box\_neg}(j, B) \in S$, for all $j<h$.
Hence, from rule (42), for all $e \in \Delta$ such that $\mathit{rank(\prj(e),j) }\in S$ (i.e., $k_\emme(e)=j<h$)
$\mathit {-inst}(\prj(e), B) \in S$ and therefore, $\mathit {inst}(\prj(e), B) \not \in S$.
Thus, for all $e \in \Delta$ such that $k_\emme(e)=j<h$, $e \not \in B^I$.
So, $d \in (\tip(B))^I$.

For $\tip(B) \sqsubseteq C$ in KB, we have $\mathit{subTyp(B,C)} \in S$.
Let $d \in (\tip(B))^I$. We have to prove that $d \in A^I$.
Assume that $k_\emme(d)=h$, i.e., $\mathit{rank(\prj(d),h) } \in S$.
As $d \in (\tip(B))^I$, $d \in B^I$ and, for all $e \in \Delta$ such that $k_\emme(e)=j<h$, $e \not \in B^I$
(and hence, by construction, $\mathit {inst}(\prj(e), B) \not \in S$).
From $d \in B^I$, by the definition of $\emme$, $\mathit {inst}(\prj(d), B) \in S$.

Consider also the rank of $aux_B$.
Let $\mathit{rank(aux_B,j) } \in S$. By rule (51) it must be that $\mathit {inst}(aux_B, B) \in S$.
Either $j=h$ or $j\neq h$.
If $j=h$, then from $\mathit{rank(aux_B,h) } \in S$, we conclude by rule (50) that
 $\mathit {box\_neg}(h, B) \in S$,
 and, given that $\mathit {inst}(\prj(d), B)$ and $\mathit{rank(\prj(d),h) }$ are in $S$, by rule (48),
 $\mathit {typ}(\prj(d), B) \in S$. Thus, by rule (31), $\mathit {inst}(\prj(d), C) \in S$.

We can exclude the case $j\neq h$, as both the hypothesis $j<h$ and the  hypothesis $j>h$ lead to a contradiction.
For $j<h$: the fact that $\mathit {inst}(aux_B, B) \in S$ contradicts the fact that, for all $e \in \Delta$ such that $k_\emme(e)=j<h$, $\mathit {inst}(\prj(e), B) \not \in S$.
For $j>h$: from $\mathit{rank(aux_B,j) } \in S$, we can conclude by (50) that $\mathit {box\_neg}(j, B) \in S$,
which would imply, by (41) and (42), that $\neg \mathit {inst}(\prj(d), B) \in S$ (from the fact that  $\mathit{rank(\prj(d),h) } \in S$  and $h<j$).
Again a contradiction.

Hence, $\emme$ is a model of KB.
For $Q= C(a)$, from the hypothesis $-inst(a,C) \in S$, hence $inst(a,C) \not \in S$ and, by construction, $a^I \not \in C^I$ in $\emme$.
For $Q= \tip(C) (a)$, from the hypothesis $-typ(a,$ $C) \in S$, hence $typ(a,C) \not \in S$.
If $\mathit {inst}(a, C) \not \in S$ then, by construction of $\emme$, $a^I \not \in C^I$ and, clearly, $a^I \not 	 \in (\tip(C))^I$.
Instead, if $\mathit {inst}(a, C) \in S$, as $\mathit{typ(a,C)} \not \in S$, it must be that, for $\mathit{rank(a,h)}$ and $\mathit{rank(aux_C,j)}$ in $S$,
$h\neq j$ (otherwise, by rules (48) and (50), would conclude $\mathit{typ(a,C)} \in S$).
Also, it can be seen that the hypothesis $h < j$ leads to a contradiction. Hence, $h>j$ and, by construction,
$k_\emme(a)> k_\emme(C)=j$, so that $a^I \not 	\in (\tip(C))^I$.

This completes the proof of Proposition 2.

\subsection{Proof of Proposition 3} \label{appendix:Prop3}

{\em Proposition 3.}
{\em For a  $\sroelrt$ knowledge base $K$ in normal form and a query $Q$, if $\emme=(\Delta, <,\cdot^I)$ is a model of $K$
falsifying a query $Q$,
then there exists an answer set $S$ of the ASP program $\Pi(K) \cup \{- \pi_Q\}$. 
}

\begin{proof}
Let $Q$ be a query $C(a)$ (respectively, $\tip(C)(a)$).
We show that such an answer set $S$ can be constructed from the model $\emme$
such that $\mathit{inst(a,C)\in S}$ (respectively, $\mathit{typ(a,C)\in S}$).
Without loss of generality, we can assume that $\emme$ has no more than $max_K+1$ different rank values
(from 0 to $max_K$)
and that the rank values have been made contiguous, according to Theorem 1.
In the ASP program we let the upper bound $n$ to be equal to $max_K$
and, in the following, we let $h_{max}$ be the maximum rank of domain elements in $\emme$ (observe that $h_{max} \leq max_K$).
We exploit $\emme$ to construct the answer set $S$ by assigning the ranks to the constants in $N_I$ and to the auxiliary constants
$\auxARC$ and $aux_C$ according to the ranks of the elements in $\emme$.

Let $S_0$ contain the following facts: 
\begin{quote}

0. $nom(c)$ for $c \in N_I$; \  $auxsupex(c)$ for $c=\auxARC$; \ $auxtc(aux_B,B)$ for all $\tip(B)$ in $K$ or $Q$;

1. $ind(c)$ for all $c \in N_I$ and for all $c$ auxiliary constants;

2. $\mathit{rank(c,h) } $, if $k_{\emme}(c^I)=h$, for each $c \in N_I$;

3. $\mathit{rank(aux_B,h) }$,
if there exists $x \in (\tip(B))^I$ and $k_{\emme}(x)=h$;

4. $\mathit{rank(aux_B,h_{max})}$ if $B^I = \emptyset$;

5. $\mathit{rank(\auxARC,h) }$ if $A^I \neq \emptyset$ and
 $h= min \{ k_{\emme}(x)\mid \; x \in (C \sqcap \exists R^-.A))^I\}$;

6. $\mathit{rank(\auxARC,h_{max})}$ if $A^I = \emptyset$;

7. $\mathit{inst(aux_B,B) } \in S$, if $B^I \neq \emptyset$, for $B \in N_C$ and $\tip(B)$ occurring in $K$;
otherwise,  let $\mathit{-inst(aux_B,B) } \in S$.

8. $\mathit{-inst(a,C) } \in S$, if $Q=C(a)$;   

9. $\mathit{-typ(a,C) } \in S$,  if $Q=\tip(C)(a)$; 

10. $\mathit{L } \in S$,  for any $L\in \Pi_K$, where $L$ is the ASP literal representing  a rule in $K$ (according to the input translation in Section 4 (Part 1)
and in \ref{Appendix_Calculus}).

11. $upperbound(max_K), poss\_rank(0), \ldots, poss\_rank(max_K), some\_at(0),\ldots,  some\_at(h_{max})$
\end{quote}
The rank of $c \in N_I$ is equal to the rank of $c^I$ in $\emme$.
The rank of $aux_B$ is equal to the rank of any typical $B$ element in $\emme$, if any (as all the typical $B$ elements have the same rank in $\emme$).
$\auxARC $ is given the rank $h_{max}$, when $A^I= \emptyset$, otherwise it is given a minimal rank of the elements in the $(C \sqcap \exists R^-.A)^I$ concept interpretation\footnote{Notice that, although inverse roles are not in the language of $\sroelrt$, at the semantic level the set of domain elements in $(C \sqcap \exists R^-.A)^I$ is well defined, according to the usual semantics of inverse roles \cite{HorrocksIGPL00}, i.e., $(\exists R^-.A)^I= \{ x \in \Delta \mid$ exists $y \in A^I$ such that $(y,x) \in R^I \}$.}.
Also, by item 5, $aux_B$ is set to be an instance of concept $B$ if and only if $B$ has some instance in $\emme$.

As in the proof of soundness of the materialization calculus in  \cite{jeliaReport} (see Lemma 2), we assign a concept expression
$\kappa(c)$ to each constant occurring in the ASP program $\Pi(K) \cup \{-\pi_Q\}$:
\begin{quote}
- if $c\in N_I$, then $\kappa(c)= \{c\}$;

- if $c=\auxARC$, then $\kappa(c)= C \sqcap  \exists R^-.A$;

- if $c=aux_B$, then $\kappa(c)=\tip(B)$.
\end{quote}

We say that a set of literals {\em $S$  is satisfied in the model $\emme$ }, if
the following conditions hold:\\
- for $B \in N_C$, if $\mathit{inst(c,B) } \in S$, then $\emme \models \kappa(c) \sqsubseteq B$ and  $\kappa(c)^I\neq \emptyset$\\
- for $d \in N_I$, if $\mathit{inst(c,d) } \in S$, then $\emme \models \kappa(c) \sqsubseteq \{d\}$ and  $\kappa(c)^I\neq \emptyset$\\
- for $B \in N_C$, if $\mathit{typ(c,B) } \in S$, then $\emme \models \kappa(c) \sqsubseteq \tip(B)$ and  $\kappa(c)^I\neq \emptyset$\\
- for $R \in N_R$, if $\mathit{triple(c,R,d) } \in S$, then $\emme \models \kappa(c) \sqsubseteq \exists R. \kappa(d)$ and  $\kappa(c)^I\neq \emptyset$\\
-  for $R \in N_R$, if $\mathit{self(c,R) } \in S$, then $\emme \models \kappa(c) \sqsubseteq \exists R. Self$ and  $\kappa(c)^I\neq \emptyset$\\
- if $\mathit{rank(c,h) } \in S$ and  $\kappa(c)^I\neq \emptyset$, then $k_{\emme}(\kappa(c))=h$\\
- if $\mathit{box\_neg(h,A) } \in S$ then, for all $x \in \Delta$ such that $k_{\emme}(x)=h$, $x \in (\Box \neg A)^I$\\
- if $\mathit{- box\_neg(h,A) } \in S$ then, for all $x \in \Delta$ s.t. $k_{\emme}(x)=h$, $x \not \in (\Box \neg A)^I$\\ 
- for $B \in N_C$, if $\mathit{-inst(c,B) } \in S$ and  $\kappa(c)^I\neq \emptyset$, then $\emme \not \models \kappa(c) \sqsubseteq B$ \\
- for $B \in N_C$, if $\mathit{-typ(c,B) } \in S$ and  $\kappa(c)^I\neq \emptyset$, then $\emme \not \models \kappa(c) \sqsubseteq \tip(B)$ \\
- for $B \in N_C$, if $\mathit{bot(B) } \in S$, then $\emme \models B \sqsubseteq \bot $\\  
- for $B \in N_C$, if $\mathit{top(B) } \in S$, then $\emme \models \top \sqsubseteq B $ 

Notice that, from the previous conditions 
it is not the case that $\mathit{bot(B) }$ and $\mathit{inst(a,B) }$ are both in $S$, for some $B \in N_C$,
otherwise, we would have (from $\mathit{inst(a,B) } \in S$) $\emme \models \kappa(a) \sqsubseteq B$ with $\kappa(a)^I\neq \emptyset$
and (from $\mathit{bot(B) } \in S$) that $\emme \models B \sqsubseteq \bot $.

Let us consider the portion $P_0$ the ASP program $\Pi(K) \cup \{- \pi_Q\}$ containing $\Pi_{K}$, 
plus the rules  (32)-(39), the rules (52), (53) and the fact $- \pi_Q$.
Once a unique rank is assigned to each constant $c$ in $N_I$ and to auxiliary constants, 
and the rank values are all contiguous and start from 0
(as required by  rules (38) and (39)), and in particular
 the rank of the typical $B$ elements (if any) have been fixed
(as in $\emme$) by introducing $\mathit{rank(aux_B,h) }$ in $S$, for some $h$, and $\mathit{inst(aux_B,B) }$
if $B^I \neq \emptyset$,
the set $S_0$ satisfies the ASP rules in $P_0$ and is supported,
that is, $S_0$ is an answer set of the program $P_0$. 

All the other rules in the program do not involve default negation and their application uniquely determines
an answer set, if it exists. 
So if there is an answer set of the ASP program $\Pi(K) \cup \{- \pi_Q\}$ it can be obtained by repeatedly applying the rules in $P_1$
containing all the rules $\Pi_{IR}$ (Part 2) 
and the rules (40)-(51), (54)
in $\Pi_{T}$ (Part 3).

We can show that the application of the rule of the program preserves the property that $S$ is satisfied in the model $\emme$.
Starting from $S_0$, which is an answer set of the portion $P_0$ of the program
we show that 
the iterative application of the remaining ASP rules (those in $P_1$) gives a new set $S$ of literals that is satisfied in $\emme$.

The proof can be done by induction on the number of applications of the rules used to add a given literal in $S$.
%

Let $S$ be the set of literals obtained after the exhaustive application of all the rules in $P_1$ starting
from $S_0$.
$S$ is satisfied by the model $\emme$ of KB. Hence, $S$ cannot contain complementary literals such as $\mathit{inst(b,A) }$
and $\mathit{-inst(b,A) }$, otherwise $S$ would not be satisfied in $\emme$.
Also, $\mathit{inst(a,C) }$ and $\mathit{bot(C)}$ cannot be in $S$ for any $a$ and $C$.
Therefore, $S$ is a consistent set of literals, and satisfies all the rules in $P_1$ as well as in $P_0$.
Moreover, any literal in $S$ is supported in $S$ because it either belongs to $S_0$ (and is supported
in $P_0$), or it is derived from $S_0$ by a sequence of rule applications.
Hence, $S$ is an answer set of $\Pi(K) \cup \{- \pi_Q\}$.
By construction, $\mathit{-inst(a,C) } \in S$ (resp., $\mathit{-typ(a,C) } \in S$). 
\end{proof}

\section{Proofs for Section 5} \label{appendix:minimal models vs AS}

\noindent
{\em Proposition 5 }

\noindent
Given a normalized knowledge base $K$ and a query $Q$,
if there is a model $\emme=(\Delta, <, \cdot^I)$ of $K$ which is $\tip$-minimal wrt $K, Q$ and falsifies  $Q$, 
then there is an answer set  $S$ of the ASP program $\Pi(K)$, which is $\tip$-minimal wrt $K, Q$ and such that $\mathit{\pi_Q \not \in S}$; and vice-versa.

\medskip
\noindent
\begin{proof}
Let $\emme=(\Delta, <, \cdot^I)$ of $K$ which is $\tip$-minimal wrt $K, Q$ and falsifies  $Q$.
By Proposition 3,
there exists an answer set $S$ of the ASP program $\Pi(K) \cup \{- \pi_Q\}$.
As $\emme$ is $\tip$-complete, by construction, $S$ is also $\tip$-complete.
Also, by construction, the ranks of the concepts $C \in {\cal T}_{K,Q}$
are the same in $\emme$ as in $S$  (i.e., $k_\emme(C)=h< \infty$ iff $\mathit{rank(aux_C,h), inst(aux_C,C) \in S}$).
We have to show that $S$ is $\tip$-minimal wrt $K, Q$.
Suppose, by absurdum, that $S$ is not $\tip$-minimal. Hence, there is a $\tip$-complete answer set $S'$ of $\Pi(K)$ such that $S' \preceq_\tip S$.
By Proposition 2, from $S'$ we can build a model $\emme'$ of $K$ such that the ranks of the concepts $C \in {\cal T}_{K,Q}$
are the same in $\emme'$ as in $S'$ (see the construction in \ref{appendix:Section4}, Section \ref{appendix:Prop2}).
By construction, $\emme'$ is also $\tip$-complete. Hence, there is a $\tip$-complete model $\emme'$ of $K$
such that $\emme' \preceq_\tip \emme$, which contradicts the hypothesis that $\emme$ is $\tip$-minimal.

Vice-versa, let $S$ be an answer set of the ASP program $\Pi(K)$, which is $\tip$-minimal wrt $K, Q$ and such that $\mathit{inst(a,C) \not \in S}$.
By Proposition 2, from $S$ we can build a model $\emme$ of $K$ such that the ranks of the concepts $C \in {\cal T}_{K,Q}$
are the same in $\emme$ as in $S$. By construction $\emme$ is $\tip$-complete (as $S$ is $\tip$-complete).
We have to show that $\emme$ is a $\tip$-minimal model of $K$. Suppose by absurdum that $\emme$ is not $\tip$-minimal.
Then, there is another $\tip$-complete model $\emme'$ of $K$ such that $\emme' \preceq_\tip \emme$.
By Proposition 3, there exists an answer set $S'$ of the ASP program $\Pi(K) \cup \{- \pi_Q\}$.
By construction, $S'$ is $\tip$-complete and assigns to the concepts $C \in {\cal T}_{K,Q}$
the same ranks as $\emme'$ (see the construction in \ref{appendix:Section4}, Section \ref{appendix:Prop3}). Hence, it must be that $S' \preceq_\tip S$, which contradicts the hypothesis that $S$ is $\tip$-minimal.
\end{proof}

\medskip

\noindent
{\em Proposition 6 }

\noindent
The problem of deciding the existence of a $\tip$ minimal answer set of $\Pi(K)$ falsifying $\pi_Q$ is in $\Sigma^P_2$.

\medskip

\begin{proof}
This problem can be solved by nondeterministically guessing a set $S$ of literals
of polynomial size in the size of $K$ and then verifying that: \\
(1) $S$ is an answer set of $\Pi(K)$;\\
(2) $S$ is $\tip$-complete wrt $K$, $Q$;\\
(3) $\pi_Q \not \in S$;\\
(4) $S$ is $\tip$-minimal wrt $K$, $Q$ among the $\tip$-complete answer sets of $\Pi(K)$.

Verification of (1), (2) and (3) requires polynomial time in the size of $K$.
In particular, for (1) the Gelfond and Lifschitz' transform  
of $\Pi(K)$ wrt $S$,   $\Pi(K)^S$
(which has polynomial size and does not contain default negation),
can be computed in polynomial time as well as its logical consequences.
For (2), $\tip$-completeness can be verified by checking if $\mathit{inst(aux_C, C)}$ is in $S$,  for all the $aux_C \in Aux_{K,Q}$
such that $\mathit{satisfiable(C)}$ holds
(using the definition of predicate $\mathit{satisfiable}$ in Section 5
based on the polynomial encoding of $K$ in \cite{DL2016+CILC}).
(4) can be checked by calling an \textsc{NP} oracle  which verifies
that $S$ is $\tip$-minimal among the $\tip$-complete answer sets of $K$.
 In fact, the verification that $S$ is not a $\tip$-minimal
answer set of $K$ can be done by an \textsc{NP} algorithm which
nondeterministically generates a set of literals $S'$ (of
polynomial size in the size of $K$) such that $S' \preceq_{\tip} S$
($S' \preceq_{\tip} S$ can be checked in polynomial time).
Hence, the problem of deciding existence of $\tip$ minimal answer set of $\Pi(K)$ falsifying $\pi_Q$ is in $NP^{NP}$.
\end{proof}


\end{appendix}

\end{document}